\documentclass{article}
\usepackage{ijcai24}

\usepackage{times}
\usepackage{soul}
\usepackage{url}
\usepackage[hidelinks]{hyperref}
\usepackage[utf8]{inputenc}
\usepackage[small]{caption}
\usepackage{graphicx}
\usepackage{amsmath}
\usepackage{amsthm}
\usepackage{booktabs}
\usepackage{algorithm}
\usepackage{algorithmic}
\usepackage[switch]{lineno}
\usepackage{graphicx}
\usepackage{subfigure}
\usepackage{amsmath}
\usepackage{amsthm}
\usepackage{bm}
\usepackage{bbm}
\usepackage{hyperref}
\usepackage{url}
\usepackage{algorithm}
\usepackage{algorithmic}
\usepackage{graphicx}
\usepackage{textcomp}
\usepackage{subfigure} 
\usepackage{multirow}
\usepackage{multicol}
\usepackage{xcolor, colortbl}
\usepackage{bm}
\usepackage{bbm}
\usepackage[normalem]{ulem}
\usepackage{enumitem}
\usepackage{caption}
\usepackage{nicefrac}

\def \mL{\mathcal L}
\newcommand{\bA}{\mathbf{A}}

\newcommand{\by}{\mathbf{y}}
\newcommand{\bh}{\mathbf{h}}

\newcommand{\bb}{\mathbf{b}}

\newcommand{\bx}{\mathbf{x}}

\newcommand{\br}{\mathbf{r}}

\newcommand{\cT}{\mathcal{T}}

\newcommand{\cV}{\mathcal{V}}
\newcommand{\cE}{\mathcal{E}}

\newcommand{\cS}{\mathcal{S}}
\newcommand{\cQ}{\mathcal{Q}}

\newcommand{\cX}{\mathcal{X}}

\newcommand{\agg}{\texttt{AGG}}
\newcommand{\update}{\texttt{UPD}}
\newcommand{\readout}{\texttt{READOUT}}

\usepackage{array}
\newcolumntype{L}[1]{>{\raggedright\let\newline\\\arraybackslash\hspace{0pt}}m{#1}}
\newcolumntype{C}[1]{>{\centering\let\newline  \\\arraybackslash\hspace{0pt}}m{#1}}
\newcolumntype{R}[1]{>{\raggedleft\let\newline \\\arraybackslash\hspace{0pt}}m{#1}}

\newcommand{\TheName}[0]{PACIA}

\title{PACIA: Parameter-Efficient Adapter for Few-Shot Molecular Property Prediction}

\author{
	Shiguang Wu$^1$
	\and
	Yaqing Wang$^2$\thanks{Corresponding author.}\and
	Quanming Yao$^1$\\
	\affiliations
	$^1$Department of Electronic Engineering, Tsinghua University\\
	$^2$Baidu Research, Baidu Inc.\\
	\emails
	wsg23@mails.tsinghua.edu.cn,
	wangyaqing01@baidu.com,
	qyaoaa@tsinghua.edu.cn
}

\begin{document}

\maketitle

\begin{abstract}
	Molecular property prediction (MPP) plays a crucial role in biomedical applications, but it often encounters challenges due to a scarcity of labeled data. 
	Existing works commonly adopt gradient-based strategy to update a large amount of parameters for task-level adaptation.  However, the increase of adaptive parameters can lead to overfitting and poor performance. 
	Observing that graph neural network (GNN) performs well as both encoder and predictor, 
	we propose PACIA, a parameter-efficient GNN adapter for few-shot MPP. 
	We design a unified adapter to generate a few adaptive parameters to modulate the message passing process of GNN. 
	We then adopt a hierarchical adaptation mechanism to 
	adapt the encoder at task-level and the predictor at query-level by the unified GNN adapter. 
	Extensive results show that \TheName{} obtains the state-of-the-art performance in few-shot MPP problems,  
	and our proposed hierarchical adaptation mechanism
	is rational and effective.
\end{abstract}

\section{Introduction}
\label{sec:intro}

Molecular property prediction~(MPP) 
which predicts whether desired properties will be active  
on given molecules, 
can be naturally modeled as a few-shot learning problem~\cite{waring2015analysis,altae2017low}. 
As wet-lab experiments
to evaluate the actual properties of molecules are expensive and risky, usually only a few labeled molecules are available for a specific property. 
While recently, 
Graph Neural Network~(GNN) is popularly used to learn molecular representations~\cite{xu2019powerful,yang2019analyzing,xiong2019pushing}. 
Modeling molecules as graphs, 
GNN can capture inherent structural information. 
Hence, 
GNN-based methods obtain better performance
than classical ones~\cite{unterthiner2014deep,ma2015deep}, especially when they are pretrained on self-learning tasks constructed from additional large scale corpus. 
As for tasks with only a few labeled molecules, 
the performance of existing GNN-based methods is still far from desired.

Various few-shot learning (FSL) methods have been developed to handle few-shot MPP problem. 
The earlier work 
IterRefLSTM~\cite{altae2017low} builds a metric-based model upon matching network \cite{vinyals2016matching}. 
Subsequent works mainly 
adopt gradient-based meta-learning strategy~\cite{finn2017model}, 
which learns parameter initialization with good generalizability across different properties 
and adapts parameters by gradient descent for target property. 
Specifically, 
Meta-MGNN~\cite{guo2021few} brings chemical prior knowledge in the form of molecular reconstruction loss, and optimizes all parameters by gradient descents. 
PAR~\cite{wang2021property} introduces attention and relation graph module to better utilize the labeled samples for property-adaptation with the awareness of target chemical property, and conducts a selective gradient-based meta-learning strategy. 
ADKF-IFT~\cite{chen2022meta} takes a gradient-based meta-learning strategy with implicit function theorem to avoid computing expensive hypergradients, and 
builds a Gaussian Process for each task as classifier. 
There are also works that bring auxiliary information such as additional reference molecules from large molecule database \cite{schimunek2023context} and auxiliary properties \cite{zhuang2023graph} 
to improve the performance of few-shot MPP \cite{schimunek2023context,zhuang2023graph}. 

\begin{figure*}[t]
\centering
\subfigure[\label{fig:ada-ratio}]{
	\includegraphics[width=0.365\textwidth]{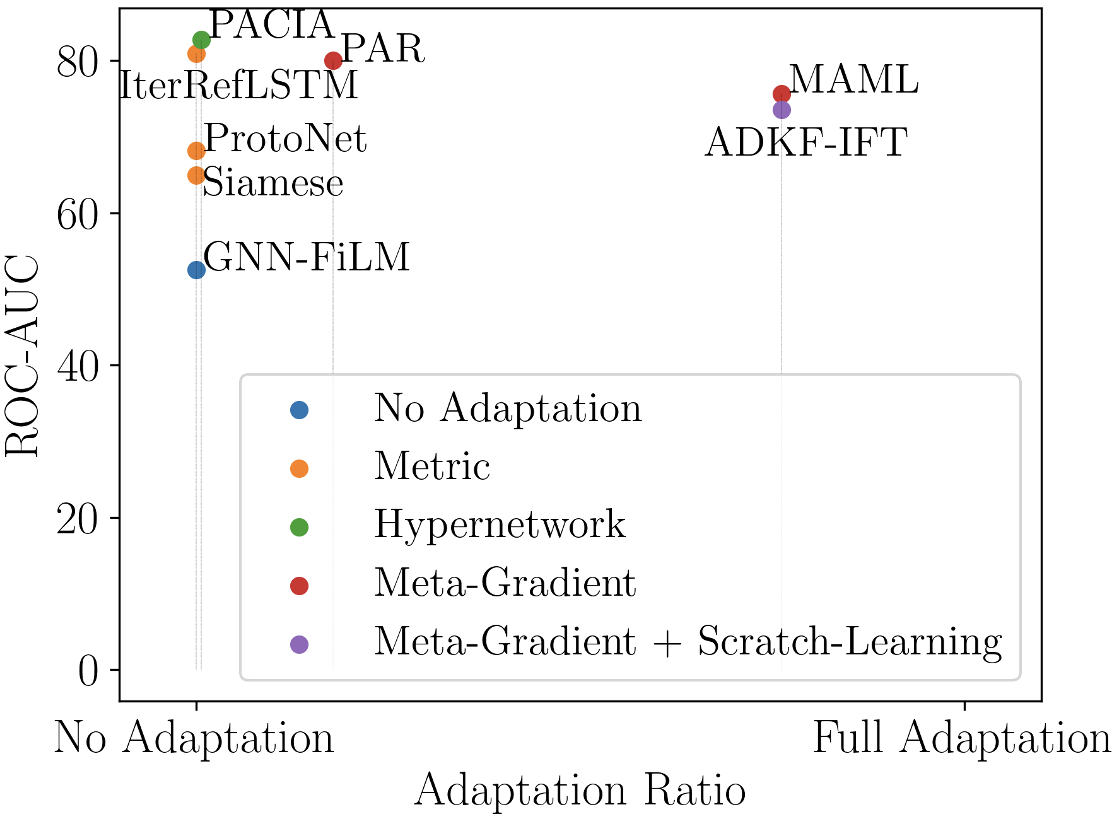}}
\subfigure[\label{fig:structure}]{
	\includegraphics[width=0.62\textwidth]{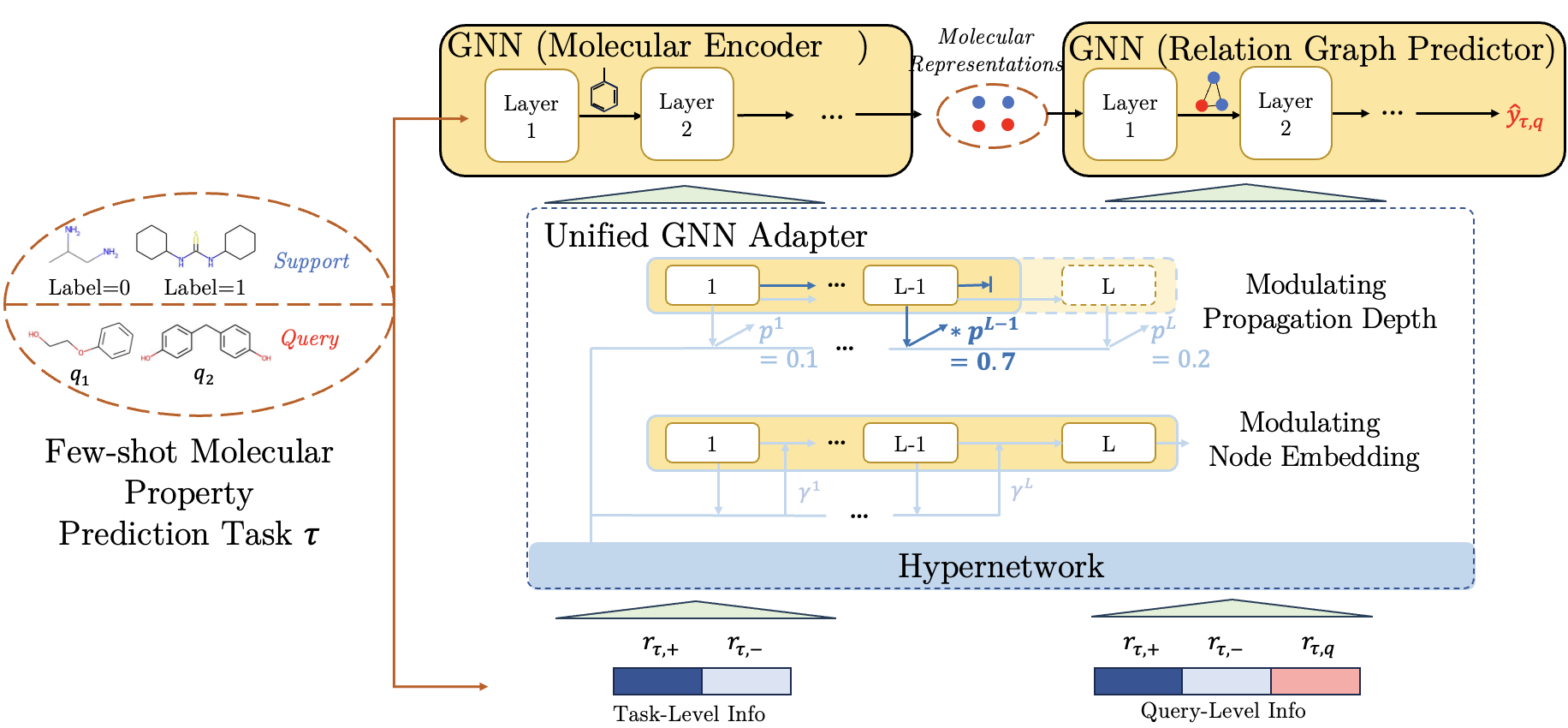}}
\caption{Illustration of the proposed \TheName{}. 
	(a). The adaptation ratio ($\frac{|adaptive~params|}{|total~params|}$) of different methods and their testing performance on 1-shot tasks of Tox21. 
	(b). Under the encoder-predictor framework, 
	\TheName{} uses a unified GNN adapter, where hypernetworks are used to generate adaptive parameters, 
	to modulate node embeddings and propagation depths in both GNN encoder and predictor. }
\label{fig:model}
\end{figure*}

Two primary issues persist in existing studies.
First, 
gradient-based meta-learning strategy 
requires updating a large number of parameters in order to adapt to each task. 
This results in poor learning efficiency and is prone to overfit given insufficient labeled samples~\cite{rajeswaran2019meta,yin2020meta}, as demonstrated in Figure~\ref{fig:model} (a). 
While with more task-specific parameters, the model gets easily overfit, which would be more severe in extreme few-shot cases. 
Effective adaptation should be made in a parameter-efficient way, i.e, without modulating a large amount of parameters. 
Second,
query-level adaptation is absent but it is important specially for few-shot MPP. 
The chemical space is enormous and the representations of molecules vary in a wide range.
The query-level difference should be addressed when classifying the encoded molecules. When query molecules are more similar to the labeled molecules in one class, 
they can be easily classified. 
While others exhibit comparable similarity to both categories, they will be harder to be  accurately classified. 
Thus, a fixed
predictor can not fit all molecules even in a single task.

In this paper, we first summarize existing works into an encoder-predictor framework, where GNN performs well acting as both encoder and predictor. 
Upon the framework, 
we propose  \textbf{\TheName{}}, a \textbf{PA}rameter-effi\textbf{CI}ent \textbf{A}dapter for few-shot MPP problem. 
To sum up, our contributions are as follows:
\begin{itemize}[leftmargin=*]
\item We propose query-level adaptation for few-shot MPP problem and design a hierarchical 
adaptation mechanism for the encoder-predictor framework generally adopted by existing approaches.

\item We design a hypernetwork-based GNN adapter to achieve parameter-efficient adaptation. 
This unified GNN adapter can generate a few adaptive parameters to modulate the message passing process  of GNN in two aspects: node embedding and propagation depth.  

\item We conduct extensive results, and show \TheName{} 
obtains the state-of-the-art performance on benchmark few-shot MPP datasets MoleculeNet~\cite{wu2018moleculenet} and FS-Mol~\cite{stanley2021fs}. 
We also closely examine and validate the  
effectiveness of our hierarchical adaptation mechanism. 
\end{itemize}

\section{Related Works}
\label{sec:rel}

\subsection{Few-Shot Learning}
Few-shot learning (FSL) aims to generalize to a task with a few labeled samples~\cite{wang2020generalizing}. 
In terms of adaptation mechanism, existing FSL methods can be classified into three main categories: 
(i) gradient-based approaches \cite{finn2017model,grant2018recasting} 
learn a model which can be generalized to new task by gradient descents,
(ii)
metric-based approaches \cite{vinyals2016matching,snell2017prototypical} 
learn to embed samples into a space where similar and dissimilar samples can be easily discriminated by a distance function, 
and 
(iii) 
amortization-based approaches \cite{requeima2019fast,lin2021task,przewiezlikowski2022hypermaml} 
use hypernetworks to map the labeled samples in the task to a few parameters to adjust the main networks to be task-specific. 
Recent works \cite{requeima2019fast} found that amortization-based approaches can reduce the risk of overfitting compared with gradient-based approaches. 
They also have faster inference speed as the adapted parameters are generated by a single forward pass without taking optimization steps. 
Besides, 
the main networks can approximate various functions in addition to distance-based ones. 

\subsection{Hypernetworks}
Hypernetworks~\cite{ha2017hypernetworks} refer to neural networks which learn to generate parameters of the main network which handles the target tasks. 
Hypernetworks have been successfully used in various applications like cold-start recommendation \cite{lin2021task} and image classification \cite{przewiezlikowski2022hypermaml}. 
Designing appropriate hypernetworks is challenging,  
requiring domain knowledge to decide 
what information to be fed into hypernetworks, 
how to adapt the main network, 
and what is the appropriate architecture of hypernetworks. 
For general GNNs, 
hypernetworks are developed to modulate weight matrix in aggregation function in message passing \cite{brockschmidt2020gnn}, 
or to 
facilitate node-specific message passing \cite{nachmani2020molecule}. 
In contrast to them, we particularly consider designing parameter-efficient modulators for GNNs used in encoder-predictor framework for few-shot MPP.

\section{Preliminaries: Few-Shot MPP}
\subsection{Problem Setup}
\label{sec:problem}

In a few-shot MPP task $\cT_{\tau}$ with respect to a specific property,
each sample $\cX_{{\tau},i}$ is a molecular graph
and its label $y_{{\tau},i} \in \{0,1\}$ records whether the molecule is active or inactive on the target property. 
Only a few labeled samples are available in $\cT_{\tau}$. 
Following 
earlier works~\cite{altae2017low,stanley2021fs,chen2022meta,schimunek2023context}, 
we model a $\cT_{\tau}$  as a $2$-way classification task $\cT_{\tau}$, 
associating with 
a support set $\cS_{\tau}=\{(\cX_{{\tau},s},y_{{\tau},s})\}_{s=1}^{N_\tau}$ 
containing 
labeled samples from active/inactive class, 
and a query set
$\cQ_{\tau}=\{( {\cX}_{{\tau},q},  {y}_{{\tau},q})\}_{q=1}^{M_\tau}$ containing $M_{\tau}$ samples
whose labels are only used for evaluation. 
We consider both 
(i) balanced support sets, i.e., $\cS_{\tau}$ contains $\frac{N_\tau}{2}$ samples per class which is consistent with the standard $N$-way $K$-shot FSL setting~\cite{altae2017low}, 
and (ii) imbalanced support  sets which exist in real-world applications~\cite{stanley2021fs}. 
Our target is to learn a model from a set of tasks $\{\cT_{\tau}\}_{j=1}^{N}$ that can generalize to new task given the few-shot support set.
Specifically, the target properties are different across tasks.

\subsection{Encoder-Predictor Framework}
\label{sec:mainnet}
In the past, 
molecules are encoded with certain properties (fingerprint vectors~\cite{rogers2010extended}) 
and fed to deep networks for prediction 
\cite{unterthiner2014deep,ma2015deep}. 
While recently, 
GNNs are popularly taken as molecular encoders~\cite{li2018adaptive,yang2019analyzing,xiong2019pushing,hu2019strategies} due to their superior performance on 
learning from of topological data. 
In either case, existing works can be summarized within an encoder-predictor framework.

Consider a molecular graph $\cX=\{\cV,\cE\}$ with node feature $\bh_v$ for each atom $v\in\cV$ and edge feature $\bb_{vu}$ for each chemical bond $e_{vu}\in\cE$ between atoms $v,u$. 
A GNN encoder maps $\cX$ to molecular representation $\br$ which is a fixed-length vector. 
At the $l$th layer, 
GNN updates atom embedding $\bh_{v}^{l}$ of $v$ as 
\begin{equation}
\label{eq:gnn-update}
\bh_{v}^{l}
\!=\!\update^{l}
\!\left(\!
\bh_{v}^{l-1},
\agg^{l} 
\left(\!
\{(\bh_{v}^{l-1}\!,\!\bh_{u}^{l-1}\!,\!\bb_{vu})|u\in\mathcal{H}(v)\}
\!\right)
\!\right),\!
\end{equation}
where
$\mathcal{H}(v)$ contains neighbors of $v$. \agg{($\cdot$)} and \update{($\cdot$)} are aggregation and updating functions respectively. 
After $L$ layers, the query-level representation $\br$ for $\cX$ is obtained as 
\begin{align}
\label{eq:gnn-readout}
\br=\readout
\left( 
\{\bh_v^{L}| v\in\cV\}
\right) , 
\end{align}
where $\readout(\cdot)$ function aggregates atom embeddings.

Then, a predictor $f(\cdot)$
assigns label for a query molecule $\cX_{{\tau},q}$ given support molecules in $\cS_{\tau}$:
\begin{align}\label{eq:linear-predictor}
\hat{\by}_{\tau,q}=f(\br_{\tau,q}|\{\br_{\tau,s}\}_{s\in\cS_{\tau}}).
\end{align}
The specific choice of $f(\cdot)$ is diverse, e.g., pair-wise similarity \cite{altae2017low}, multi-layer perceptron (MLP)  \cite{guo2021few,wang2021property} and Mahalanobis distance \cite{stanley2021fs}.
Recently, GNN-based predictor which operates on relation graphs of molecules is found to 
effectively compensate for the lack of supervised information \cite{wang2021property}. 
In particular, 
molecular representations 
are refined on relation graphs
such that the similar molecules cluster closer. 
Initialize molecular representations as the output of the encoder, 
i.e., $\bh_{\tau,i}^{0}=\br_{\tau,i}$.
Denote the set of $N_\tau+1$ molecules as $\mathcal{R}_{{\tau}, q} = ( {\cX}_{{\tau},q},  {y}_{{\tau},q}) \cup \cS_{\tau}$, which contains all information to make prediction for the query molecule  ${\cX}_{{\tau},q}$. 
The relation graph works by recurrently estimating the adjacency matrix and updating the molecular representations.
At the $l$th layer,  each element $a_{ij}^l$ in
the adjacent matrix $\bA_{{\tau}, q}^{l}$ of the relation graph is learned to represent pair-wise similarities between any two molecules in $\mathcal{R}_{{\tau}, q}$:
\begin{align}
\label{eq:adj}
a_{ij}^l
=
\begin{cases}
	\texttt{MLP}
	\left( 
	|\bh_{{\tau},i}^{l-1}-\bh_{{\tau},j}^{l-1}|
	\right)
	& 
	\text{if\;} i\neq j
	\\
	1
	&
	\text{otherwise}
\end{cases}.
\end{align}
Then, 
each molecular representation is refined as
%
\begin{align}
\label{eq:node}
\bh_{{\tau},i}^l=\texttt{MLP}(\sum\nolimits_{j=1}^{N_\tau+1}a_{ij}^l \bh_{{\tau},j}^{l-1}).
\end{align}
After $L$ layers of refinement,
$\bh_{{\tau},q}^L$ and  $\bh_{{\tau},s}^L$ (in place of $\br_{\tau,q}$ and $\br_{\tau,s}$)
are fed to \eqref{eq:linear-predictor} to obtain final prediction
$\hat{\bm{y}}_{{\tau},q}$ for $\cX_{\tau}$. 

\section{Hierarchical Adaptation of Encoder-Predictor Framework}
To generalize across different tasks with a few labeled molecules, 
existing works \cite{wang2021property,chen2022meta} usually conduct
task-level adaptation by gradient-based meta-learning.
(see Appendix~\ref{app:maml}). 
However, as discussed in Section~\ref{sec:rel}, 
gradient-based meta-learning optimizes most parameters using a few labeled molecules, 
which is slow to optimize and easy to overfit. 
As for query-level adaptation, gradient is not accessible for each query molecule in testing since the label is unknown during training.
Therefore, 
we turn to hypernetworks to achieve parameter-efficient adaptation. 
Task-level adaptation is achieved in encoder since the structural features in molecular graphs needs to be captured  in a property-adaptive manner, while query-level adaptation is achieved in predictor based on the property-adaptive representations. 

As discussed in Section~\ref{sec:mainnet}, GNNs can effectively act as both 
encoder and predictor. 
Therefore, 
we propose
\TheName{} (Figure \ref{fig:model}), a method using a unified GNN adapter  
to generate a few adaptive parameters to 
hierarchically 
adapt the encoder at task-level and the predictor at query-level in a parameter-efficient manner.  
In the sequel, we provide the details of the unified GNN adapter (Section~\ref{sec:proposed-gnn-adapter}), 
then describe how to learn the main networks (including encoder and predictor)
with the unified GNN adapter by episodic training (Section~\ref{sec:train}). 
Finally, we present a comparative discussion of \TheName{} in relation to existing works (Section~\ref{sec:discussion}). 

\subsection{A Unified GNN Adapter}
\label{sec:proposed-gnn-adapter}

To adapt GNN's parameter-efficiently, 
we design a GNN adapter to modulate the node embedding and propagation depth, 
which are essential in message passing process. 

\paragraph{Modulating Node Embedding.} 
Denote the node embedding
at the $l$th layer as $\bh^l$, which can be atom embedding in encoder or 
molecular embedding in relation graph predictor. 
We obtain 
adapted embedding $\hat{\bh}^{l}$ as
\begin{align}\label{eq:film}
\hat{\bh}^{l}
=e(\bh^l,\bm{\gamma}^l),
\end{align}
where $e(\cdot)$ is an element-wise function, and $\bm{\gamma}^l$ is adaptive
parameter generated by the hypernetworks. 
This adapted embedding $\hat{\bh}^{l}$ is then fed to next layer of 
GNN.

\paragraph{Modulating Propagation Depth.}
Further, 
we manage to modulate the propagation depth, i.e., layer number $l$ of a GNN. 
Controlling $l$ is challenging since it is discrete. 
We achieve this by training a differentiable controller, which is a hypernetwork to generate a scalar $p^{l}$ corresponding to the $l$th layer.
The value of $p^{l}$ estimates how likely the message passing should stop after the $l$th layer.
Finally,
assume there are $L$ layers in total,
the vector
\begin{align}
\label{eq:mod-psoft}
\bm{p}=\texttt{softmax}([~p^1,~p^2,~\cdots,~p^L~]),
\end{align}
represents 
the plausibility of choosing each layer. The hypernetwork is shared across all $L$ layers. 
During meta-training, 
$\bm{p}$ is used as differentiable weight to modulate GNN layers.
Specifically, after propagation through all $L$ layers, 
the final node embedding $\widetilde{\bh}$ is obtained as 
\begin{align}
\label{eq:weightednode}
\widetilde{\bh}=\sum\nolimits_{l=1}^{L}[\bm{p}]_l \bh^l,
\end{align}
where $[\bm{p}]_l$ is the $l$th element of $\bm{p}$. 

\paragraph{Generating Adaptive Parameters.}
We generate adaptive parameters $\{\bm{\gamma}^l,p^l\}_{l=1}^L$ by hypernetworks. 
In particular, note that the generated adaptive parameter should be permutation-invariant to the order of input samples in $\mathcal{S}_{\tau}$. 
Therefore, we first calculate class prototypes $\bm{r}_{{\tau},+}^l$ and $\bm{r}_{{\tau},-}^l$  of active class (+) and inactive class (-) for samples in $\cS_{\tau}$ by 
\begin{equation}	\label{eq:mean-rep}
	\begin{aligned}
		&\!\bm{r}_{{\tau},+}^l
		\!=\!
		\frac{1}{ |\cS_{\tau}^+| | \cV_{{\tau},s} |}
		\!\sum\nolimits_{\cX_{{\tau},s} \in \cS_{\tau}^+}
		\!\texttt{MLP}(\!\big[
		\!\sum\nolimits_{v\in\cX_{{\tau},s}} \!\bh_v^{l}
		|
		\bm{y}_{{\tau},s}
		\big]\!), \\
		&\!\bm{r}_{{\tau},-}^l
		\!=\!
		\frac{1}{ |\cS_{\tau}^-| | \cV_{{\tau},s} |}
		\!\sum\nolimits_{\cX_{{\tau},s} \in \cS_{\tau}^-}
		\!\texttt{MLP}(\!\big[
		\!\sum\nolimits_{v\in\cX_{{\tau},s}} \!\bh_v^{l}
		|
		\bm{y}_{{\tau},s}
		\big]\!),
	\end{aligned}
\end{equation}
where $[\cdot |\cdot]$ means concatenating, 
$\cS_{\tau}^+$ and $\cS_{\tau}^-$ are the sets of active and inactive samples in $\cS_{\tau}$, 
and $\bm{y}_{{\tau},s}$ is the one-hot encoding of label. 
Using $\bm{r}_{{\tau},+}^l$ and $\bm{r}_{{\tau},-}^l$ allows subsequent steps to keep supervised information while being permutation-invariant. 

Recall that task-level adaptation and query-level adaptation is achieved in encoder and predictor respectively.
For task-level adaptation, 
we then map 
$\bm{r}_{{\tau},+}^l$ and $\bm{r}_{{\tau},-}^l$
to $\bm{\gamma}_\tau^l,p^l_{{\tau}}$
as 
\begin{align}
\label{eq:ada-pt}
[\bm{\gamma}_\tau^l,p^l_\tau]
= \texttt{MLP}
\left( 
\left[ 
\bm{r}_{{\tau},+}^l
~|~
\bm{r}_{{\tau},-}^l
\right]
\right).
\end{align}
As for query-level adaptation,
information comes from both $\cS_{\tau}$ and the specific query molecule $\cX_{{\tau},q}$. 
Likewise, we use class prototypes $\bm{r}_{{\tau},+}^l$ and $\bm{r}_{{\tau},-}^l$ to keep permutation-invariant. 
We then generate $\bm{\gamma}_{\tau,q}^l,p^l_{{\tau},q}$ as
\begin{align}
\label{eq:ada-pq}
[\bm{\gamma}_{\tau,q}^l,p^l_{\tau,q}]
= \texttt{MLP}
\Big( 
\big[ 
\br_{{\tau},+}^l
~|~
\br_{{\tau},-}^l
~|~
\sum\nolimits_{v\in\cX_{{\tau},q}} \bh_v^{l}
\big]
\Big),
\end{align}
combining information in $\cX_{{\tau},q}$ and $\cS_{\tau}$.
Note that parameters of these $\texttt{MLP}$s in hypernetworks are jointly meta-learned with the encoder and predictor.

\subsection{Learning and Inference}\label{sec:train}


Denote the collection of all model parameters in main network(\eqref{eq:gnn-update}-\eqref{eq:node}) and hypernetwork (\eqref{eq:mean-rep}-\eqref{eq:ada-pq}) as $\bm{\Theta}$, excluding adaptive parameters. 
Our objective takes the form: 
\begin{equation}
\label{eq:loss}
\!\min\!
\sum\nolimits_{\tau=1}^{N}\!\mL_{\tau}, \text{with}~
\mL_{\tau}
\!=\! 
\!-\! 
\sum\nolimits_{\bm{x}_{{\tau},q}\in \cQ_{\tau}}
\!\bm{y}^\top_{{\tau},q}
\log
\left( 
\hat{\bm{y}}_{{\tau},q}
\right).\! 
\end{equation}
$\mL_{\tau}$ is the loss in task $\cT_{\tau}$, 
${\bm{y}}_{{\tau},q}$ is one-hot ground-truth label vector and 
$\hat{\bm{y}}_{{\tau},q}$  is prediction obtained by \eqref{eq:linear-predictor}. 

Note that
$\bm{\Theta}$ is shared across all tasks. While the adaptive parameter $\{\bm{\gamma}^l\}_{l=1}^L$ 
in \eqref{eq:film}
and $\{p^l\}_{l=1}^L$  in \eqref{eq:mod-psoft}
are generated by 
hypernetworks. 
The size of adaptive parameter is far smaller than the main network. This realizes parameter-efficient adaptation and mitigates the risk of overfitting. 

	\begin{algorithm}[ht]
		\caption{Meta-training procedure of \TheName{}.}
		\begin{algorithmic}[1]
			\REQUIRE meta-training task set $\bm{\cT}_{\text{train}}$; 
			\STATE initialize $\bm{\Theta}$ randomly or use a pretrained one;
			\WHILE {not done}
			\FOR {each task $\cT_{\tau} \in \bm{\cT}_{\text{train}}$}
			\FOR {$l\in\{1,2,\cdots,L_{\text{enc}}\}$}
			\STATE + generate $[\bm{\gamma}^l_\tau,p^l_\tau]$ by \eqref{eq:ada-pt};
			\STATE modulate atom embedding ${\bh}^{l}_v\leftarrow e(\bh^l_v,\bm{\gamma}^l_\tau)$; 
			\STATE * update atom embedding $\bh_{v}^{l}$ by \eqref{eq:gnn-update};
			\ENDFOR
			\STATE obtain atom embedding after message passing $\bh^{L_{\text{enc}}}_v\leftarrow \sum\nolimits_{l=1}^{L_{\text{enc}}}[\bm{p}_\tau]_l \bh_v^l$ and obtain molecular embeddings by \eqref{eq:gnn-readout};
			\ENDFOR
			\FOR {each query $( {\cX}_{{\tau},q},  {y}_{{\tau},q})\in\cQ_{\tau}$}
			\FOR {$l\in\{1,2,\cdots,L_{\text{rel}}\}$}
			\STATE + generate $[\bm{\gamma}^l_{\tau,q},p^l_{\tau,q}]$s by \eqref{eq:ada-pq};
			\STATE modulate molecular embedding $\bh_{\tau,i}^{l}\leftarrow e(\bh_{\tau,i}^{l}\bm{\gamma}^l_{\tau,q})$; 
			\STATE * update molecular embedding by \eqref{eq:adj}-\eqref{eq:node};
			\ENDFOR
			\STATE obtain molecular embedding after message passing $\bh^{L_{\text{rel}}}_{\tau,i}\leftarrow \sum\nolimits_{l=1}^{L_{\text{rel}}}[\bm{p}_{\tau,q}]_l \bh_{\tau,i}^l$; 
			\STATE obtain prediction  $\hat{\by}_{\tau,q}$ by \eqref{eq:linear-predictor};
			\ENDFOR
			\STATE calculate loss by \eqref{eq:loss}; 
			\STATE update $\bm{\Theta}$ by gradient descent;
			\ENDWHILE
			\RETURN learned $\bm{\Theta}^*$.
		\end{algorithmic}
		\label{alg:train}
	\end{algorithm}
Algorithm ~\ref{alg:train}\footnote{In Algorithm \ref{alg:train} and \ref{alg:test},``*'' (resp. ``+'') indicates the step is executed by the main network (resp. hypernetwork).} 
summarizes the training procedure of \TheName{}. 
As mentioned above, our unified GNN adapter can simultaneously modulate the node embedding and propagation depth, and be cascaded to adapt both encoder and predictor. 
During training, molecular graph $\cX_{{\tau},i}$ is first processed by encoder (line 4-9). At each layer, adaptive parameters $[\bm{\gamma}_\tau^l,p^l_\tau]$ are obtained by \eqref{eq:ada-pt} (line 5).  Then, \eqref{eq:film} modulates all atom embeddings $\bh_v^l$ (line 6). 
After $L_{\text{enc}}$ layers of message passing \eqref{eq:gnn-update}, 
\eqref{eq:weightednode} is applied before \eqref{eq:gnn-readout}, to get property-adaptive molecular representations and initialize node embeddings $\bh_{{\tau},i}^0=\br_{\tau,i}$ in relation graph (line 9).
Then in predictor, at each layer of GNN on relation graph, adaptive parameters $	[\bm{\gamma}_{\tau,q}^l,p^l_{\tau,q}]$ are obtained with \eqref{eq:ada-pq} (line 13) and \eqref{eq:film} modulates all node embeddings $\bh_{\tau,i}^l$ (line 14). 
After $L_{\text{rel}}$ layers of message passing by \eqref{eq:adj}\eqref{eq:node}, \eqref{eq:weightednode} is applied (line 17). The final prediction $\hat{\by}_{\tau,q}$ is obtained by \eqref{eq:linear-predictor}. 

Testing procedure is provided in Appendix~\ref{app:method}. 
The process is similar.  
A noteworthy difference is the propagation depth is adapted by selecting the layer with maximal plausibility:
\begin{align}\label{eq:select}
l'= \operatorname{argmax}_{l\in \{1,2,\cdots,L\}}{~p^{l}}.
\end{align}
Only $\bh^{l'}$, rather than \eqref{eq:weightednode}, is fed forward to the next module.

	\begin{table*}[ht]
	\centering
	\setlength\tabcolsep{2.0pt}
	\begin{tabular}{c|cc|cc|cc|cc}
		\hline
		\multirow{2}{*}{Method}  &\multicolumn{2}{c|}{Tox21} &\multicolumn{2}{c|}{SIDER}
		& \multicolumn{2}{c|}{MUV}&\multicolumn{2}{c}{ToxCast} \\
		&10-shot&1-shot&10-shot&1-shot&10-shot&1-shot&10-shot&1-shot\\\hline
		GNN-ST & $61.23_{(0.89)}$&$55.49_{(2.31)}$&$56.25_{(1.50)}$&$52.98_{(2.12)}$&$54.26_{(3.61)}$&$51.42_{(5.11)}$&$55.66_{(1.47)}$&$51.80_{(1.99)}$\\
		MAT & $64.84_{(0.93)}$&$54.90_{(1.89)}$&$57.45_{(1.26)}$&$52.97_{(3.00)}$&$56.19_{(2.88)}$&$52.01_{(4.05)}$&$58.50_{(1.62)}$&$52.41_{(2.34)}$\\
		GNN-MT & $69.56_{(1.10)}$&$62.08_{(1.25)}$&$60.97_{(1.02)}$&$55.39_{(1.83)}$&$66.24_{(2.40)}$&$60.78_{(2.91)}$&$65.72_{(1.19)}$&$62.38_{(1.67)}$\\
		ProtoNet&$72.99_{(0.56)}$&$68.22_{(0.46)}$&$61.34_{(1.08)}$&$57.41_{(0.76)}$&$68.92_{(1.64)}$&$64.81_{(1.95)}$&$65.29_{(0.82)}$&$63.73_{(1.18)}$\\
		MAML&$79.59_{(0.33)}$&$75.63_{(0.18)}$&$70.49_{(0.54)}$&$68.63_{(1.51)}$&$68.38_{(1.27)}$&$65.82_{(2.49)}$&$68.43_{(1.85)}$&$66.75_{(1.62)}$\\
		Siamese&$80.40_{(0.29)}$&$65.00_{(11.69)}$&$71.10_{(1.68)}$&$51.43_{(2.83)}$&$59.96_{(3.56)}$&$50.00_{(0.19)}$&-&-\\
		EGNN&$80.11_{(0.31)}$&$75.71_{(0.21)}$&$71.24_{(0.37)}$&$66.36_{(0.29)}$&$68.84_{(1.35)}$&$62.72_{(1.97)}$&$66.42_{(0.77)}$&$63.98_{(1.20)}$\\
		IterRefLSTM&$81.10_{(0.10)}$&$80.97_{(0.06)}$&$69.63_{(0.16)}$&$71.73_{(0.06)}$&$49.56_{(2.32)}$&$48.54_{(1.48)}$&-&-\\
		PAR&$82.13_{(0.26)}$&$\underline{80.02}_{(0.30)}$&$\underline{75.15}_{(0.35)}$&$\underline{72.33}_{(0.47)}$&$68.08_{(2.23)}$&$65.62_{(3.49)}$&$70.01_{(0.85)}$&$\underline{68.22}_{(1.34)}$\\
		ADKF-IFT&$\underline{82.43}_{(0.60)}$&$77.94_{(0.91)}$&$67.72_{(1.21)}$&$58.69_{(1.44)}$&$\bm{98.18}_{(3.05)}$&$\underline{67.04}_{(4.86)}$&$\underline{72.07}_{(0.81)}$&$67.50_{(1.23)}$\\
		\TheName{}&$\bm{84.25_{(0.31)}}$&$\bm{82.77_{(0.15)}}$&$\bm{82.40_{(0.26)}}$&$\bm{77.72_{(0.34)}}$&$\underline{72.58_{(2.23)}}$&$\bm{68.80}_{(4.01)}$&$\bm{72.38_{(0.96)}}$&$\bm{69.89_{(1.17)}}$
		\\\hline
	\end{tabular}
	\caption{Test ROC-AUC (\%) obtained on MoleculeNet. 
		The best results 
		are bolded, second-best results are underlined.  }
	\label{tab:results}
\end{table*}

\subsection{Comparison with Existing Works}
\label{sec:discussion}
From the perspective of hypernetworks, the usage of hypernetworks for encoder is related to GNN-FiLM~\cite{brockschmidt2020gnn}, which considers a GNN as main network. It builds hypernetworks with target node as input to generate parameters of FiLM layers, to equip different nodes with different aggregation functions in the GNN. 
What and how to adapt are similar to ours, but it is different that the input of our hypernetworks for encoder is $\cS_{\tau}$ and how we encode a set of labeled graphs. 

As for few-shot learning, some recent studies~\cite{requeima2019fast,lin2021task} use hypernetworks to transform the support set into parameters that modulate the main network. The functionality of their hypernetworks is akin to that of our hypernetwork for the encoder.
However, there is a distinct difference in the architecture of main networks: while their approaches employ convolutional neural networks and MLPs, our method uniquely modulates the message passing process of GNN.
Furthermore, the application of hypernetworks for the predictor, aimed at adjusting the model architecture based on a query sample (without label information) and a set of labeled samples, has not yet been explored in the literature. 
A detailed comparison of \TheName{} w.r.t. existing few-shot MPP works is in Appendix~\ref{app:compare}.

\section{Experiments}
\label{sec: exp}

In this section, we evaluate the proposed \TheName{}\footnote{Code is available at \url{https://github.com/LARS-research/PACIA}.} on few-shot MPP problems. 
We run all experiments with 10 random seeds, 
and report the mean and standard deviations (in the subscript bracket). 
Appendix \ref{app:exp-detail} provides 
more information of datasets, baselines, and implementation details.

\subsection{Performance Comparison on MoleculeNet}

We use
Tox21~\cite{Tox21}, SIDER~\cite{kuhn2016sider}, MUV~\cite{rohrer2009maximum} and ToxCast~\cite{richard2016toxcast} 
from MoleculeNet \cite{wu2018moleculenet}, 
which are commonly used to evaluate the performance on few-shot MPP~\cite{altae2017low,wang2021property}. 
We adopt the public data split provided by  \cite{wang2021property}. 
The support sets are balanced, each of them contains $K$ labeled molecules per class, where $K=1$ and $K=10$ are considered.
The performance is evaluated by ROC-AUC 
calculated on the query set of each meta-testing task and averaged across all meta-testing tasks.

We compare \TheName{} with the following baselines:  
1) single-task method  
\textbf{GNN-ST} \cite{gilmer2017neural}; 
2) multi-task pretraining method  
\textbf{GNN-MT }\cite{corso2020principal,gilmer2017neural}; 
3) self-supervised pretraining method  
\textbf{MAT} \cite{maziarka2020molecule}; 
4) meta-learning methods, including \textbf{Siamese}~\cite{koch2015siamese}, \textbf{ProtoNet} \cite{snell2017prototypical}, \textbf{MAML} \cite{finn2017model}, \textbf{EGNN}~\cite{kim2019edge}; 
and 
5) methods proposed for few-shot MPP, including \textbf{IterRefLSTM}~\cite{altae2017low}, \textbf{PAR}~\cite{wang2021property} and \textbf{ADKF-IFT}~\cite{chen2022meta}. 
Note that 
MHNfs \cite{schimunek2023context} is not included as it uses additional reference molecules from external datasets, which leads to unfair comparison.
GS-META \cite{zhuang2023graph} has not been compared since that approach requires multiple properties of each molecule, which is not applicable when a molecule is only evaluated w.r.t. one property.
Following earlier works~\cite{guo2021few,wang2021property}, 
we use
GIN~\cite{xu2019powerful} as encoder, which is trained from scratch.  

Table~\ref{tab:results} shows the results. 
Results of Siamese and IterRefLSTM are copied from~\cite{altae2017low} as their codes are unavailable, and their results on ToxCast are unknown. 
GNN-FiLM is a general GNN whose target is not few-shot MPP, which explains its bad performance. 
\TheName{} obtains the highest ROC-AUC scores on all cases except the 10-shot case on MUV, 
where ADKF-IFT outperforms the others by a large margin. 
This can be a special case where ADKF-IFT works well but may not be generalizable. 
Moreover, depending on the number of local-update steps of ADKF-IFT, \TheName{} is about 5 times faster than ADKF-IFT (both meta-training and inference time is about 1/5).
In terms of average performance, 
\TheName{} significantly outperforms the second-best method 
ADKF-IFT by 3.25\%. 
We also provide results obtained with a pretrained encoder 
in Appendix~\ref{app:performance-pretrain}. 
Similar observations can be made: our  \TheName{} with pretrained encoder (Pre-\TheName{}) performs the best, and its performance gain is more significant when fewer labeled samples are provided. 
Further, results in Appendix \ref{app:more-shot} shows the performance comparison between \TheName{} and a fine-tuned GNN with varying support set size. 
This shows that 
PACIA nicely achieves its goal: handling few-shot MPP problem in a parameter-efficient way.


\subsection{Performance Comparison on FS-Mol}
\label{sec:fsmol}

We also use FS-Mol  \cite{stanley2021fs}, a new benchmark consisting of a large number of diverse tasks for model pretraining 
and a set of few-shot tasks with imbalanced classes. 
We adopt the public data split \cite{stanley2021fs}. 
Each support set contains 64 labeled molecules, and
can be imbalanced 
where the number of labeled molecules from active and inactive is not equal. 
All remaining molecules  in the task  form the query set. 
Testing tasks are divided into categories with support size 16~\cite{schimunek2023context}, which is close to real-world scenario. 
The performance is 
evaluated by $\Delta$AUPRC (change in area under the precision-recall curve) w.r.t. a random classifier~\cite{stanley2021fs}, averaged across meta-testing tasks.

We use the same baselines that were applied to MoleculeNet.  
Table~\ref{tab:fsmol} shows the results. 
We find that \TheName{} performs the best. 
Besides, the time-efficiency of \TheName{} is much higher since its adaptation only needs a single forward pass. 
While IterRefLSTM and ADKF-IFT  take multiple local-update steps, they are much slower to generalize.  

\begin{table}[htbp]
	\centering
	\setlength\tabcolsep{1.2pt}
	\begin{tabular}{c|c|c|c|c}
		\hline
		\multirow{2}{*}{Method} &All &Kinases &Hydrolases&Oxido-\\
		&[157]&[125]&[20]&reductases[7]\\\hline
		GNN-ST&$2.9_{(0.4)}$&$2.7_{(0.4)}$&$4.0_{(1.8)}$&$2.0_{(1.6)}$\\
		MAT&$5.2_{(0.5)}$&$4.3_{(0.5)}$&$9.5_{(1.9)}$&$6.2_{(2.4)}$\\
		GNN-MT&$9.3_{(0.6)}$&$9.3_{(0.6)}$&$10.8_{(2.5)}$&$5.3_{(1.8)}$\\
		MAML&$15.9_{(0.9)}$&$17.7_{(0.9)}$&$10.5_{(2.4)}$&$5.4_{(2.8)}$ \\
		PAR& $16.4_{(0.8)}$ & $18.2_{(0.9)}$ & $10.9_{(2.0)}$ & $3.9_{(0.8)}$ \\
		ProtoNet&$20.7_{(0.8)}$&$21.5_{(0.9)}$&$20.9 _{(3.0)}$&$9.5_{(2.9)}$\\
		EGNN &$21.2_{(1.1)}$&$22.4_{(1.0)}$ &$20.5_{(2.4)}$&$9.7_{(2.2)}$\\
		Siamese &$22.3_{(1.0)}$ & $24.1_{(1.0)}$ & $17.8_{(2.6)}$ & $8.2_{(2.5)}$\\
		IterRefLSTM &$\underline{23.4}_{(1.0)}$ & $\textbf{25.1}_{(1.0)}$ & $19.9_{(2.6)}$ & $9.8_{(2.7)}$\\
		ADKF-IFT&$\underline{23.4} _{(0.9)}$&${24.8} _{(2.0)}$&${\underline{21.7 }}_{(1.7)}$&$\textbf{10.6} _{(0.8)}$\\
		\TheName{}&$\textbf{23.6}_{(0.8)}$&$\textbf{25.1} _{(1.6)}$&$\textbf{21.9}_{( 2.9)}$&$\textbf{10.6} _{(1.0)}$\\\hline
	\end{tabular}
	\caption{Test $\Delta$AUPRC (\%) obtained on FS-Mol. Tasks are categorized by target protein type. The number of tasks per category is reported in brackets. The best results are bolded, second-best results are underlined.}
	\label{tab:fsmol}
	\vspace{-10px}
\end{table}

\subsection{Ablation Study}
\label{sec:ablation}

We consider various variants of \TheName{}, including 
(i) \textbf{fine-tuning}:  using the same model structure and fine-tuning all parameters to adapt to each property without hypernetworks; 
(ii) \textbf{w/o T}: removing task-level adaptation,  
thus the GNN encoder will not be adapted by  
hypernetworks w.r.t. each property; 
and 
(iii) \textbf{w/o Q}: removing query-level adaptation, 
such that all molecules are processed by the same predictor.

\begin{figure}[H]
	\centering
	\subfigure[Adaptation strategies.\label{fig:ablation1}]{
		\includegraphics[width=0.252\textwidth]{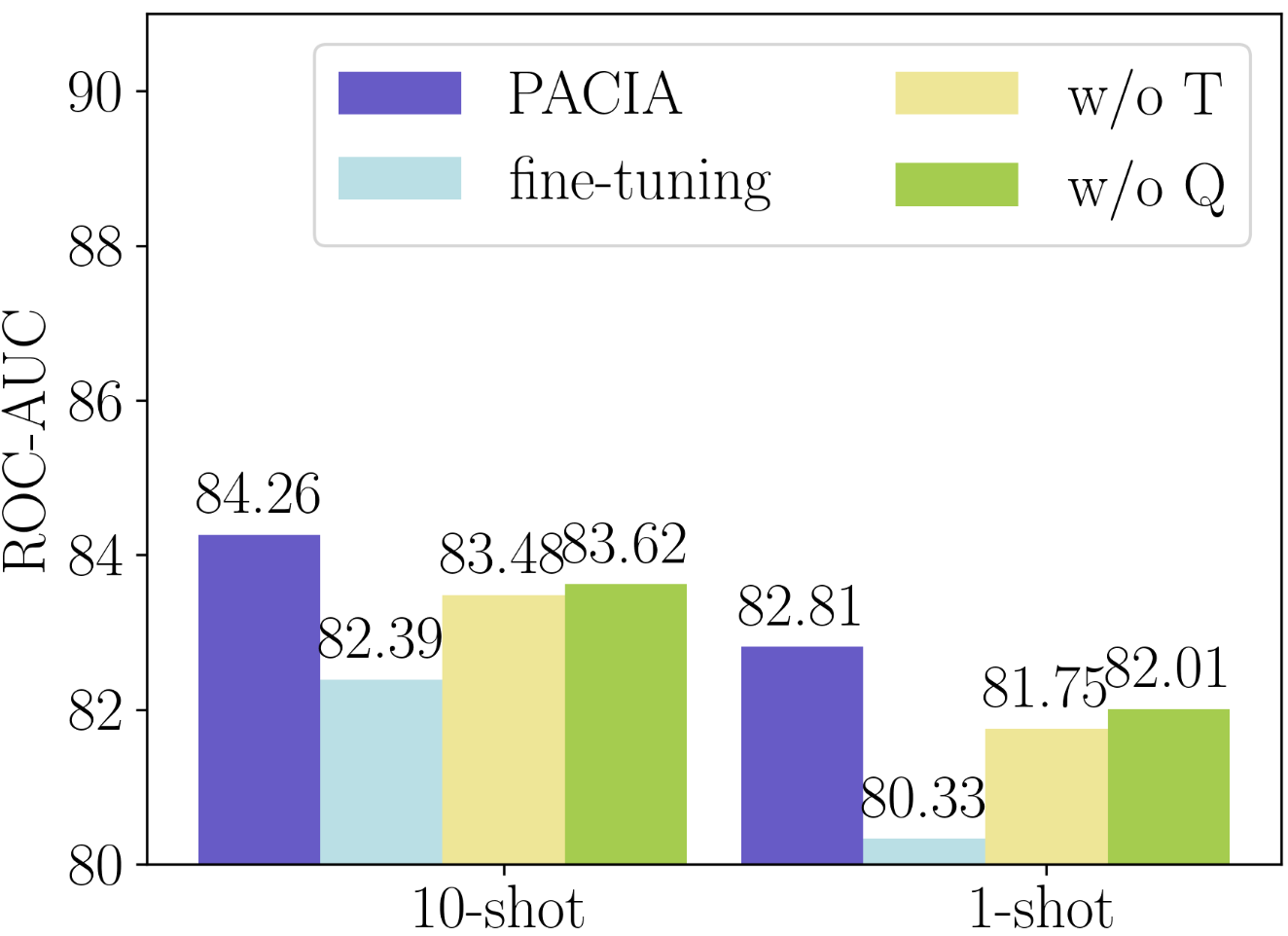}}
	\subfigure[Modulation functions.\label{fig:ablation2}]{
		\includegraphics[width=0.215\textwidth]{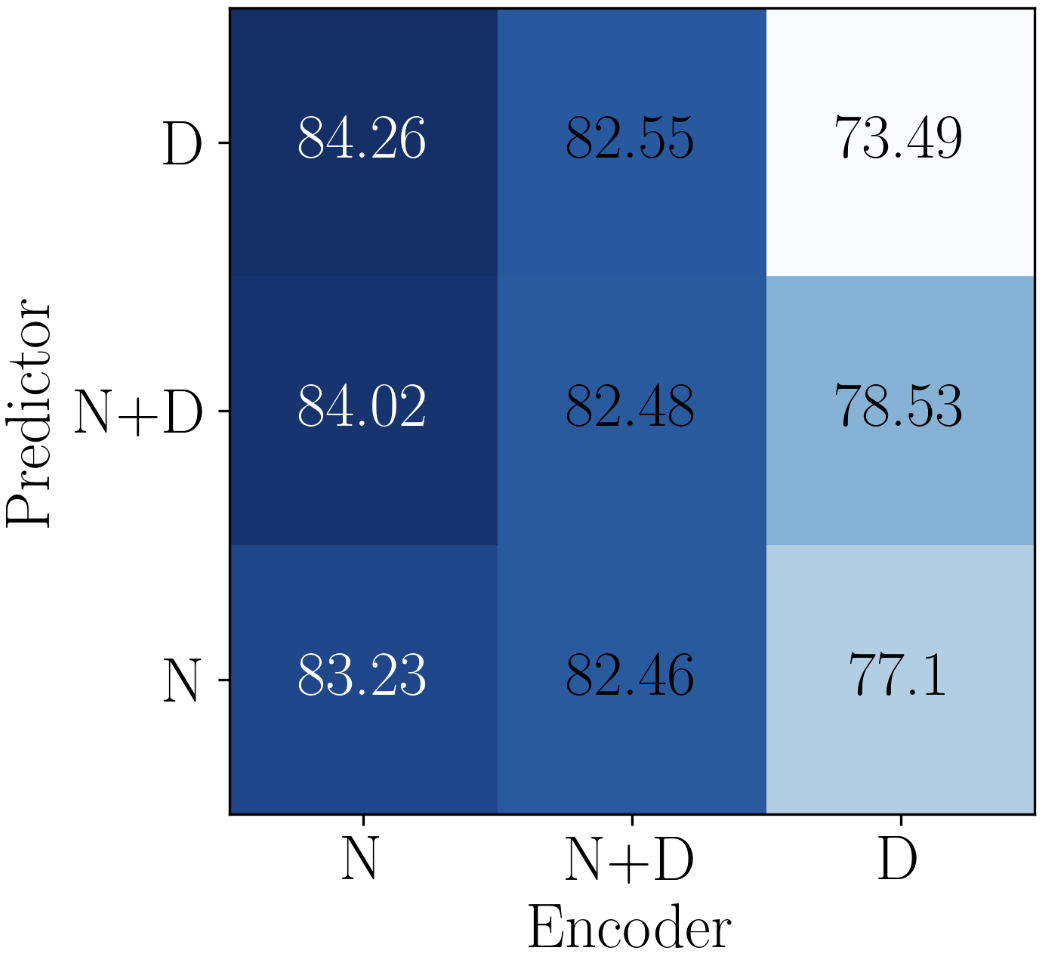}}
	\caption{Ablation study on 10-shot tasks of Tox21.}
\end{figure}
Figure~\ref{fig:ablation1} provides performance comparison on Tox21.  
As shown, the performance gain of \TheName{} over ``w/o Q" shows the necessity of query-level adaptation.  
The gap between \TheName{} and ``w/o T" indicates the effect of adapting the model to be task-specific. 
One can also notice that without query-level adaptation, ``w/o Q" still obtains better performance than gradient-based baselines like PAR, which indicates the advantage of designing the amortization-based hypernetwork. 
The poor performance of ``fine-tuning" is possibly because of the overfitting caused by updating all parameters with only a few samples. 
In sum, every component  of \TheName{} is important for achieving good performance.

Now that the effectiveness of task-level and query-level adaptation are validated, 
we further investigate modulation functions, i.e, modulating node embedding (\textbf{N}), modulating propagation depth (\textbf{D}), modulating both (\textbf{ND}), for encoder and predictor. 
There are $3\times 3$ combinations, whose performance is reported in Figure \ref{fig:ablation2}. 
We find that only modulating node embedding in encoder while only modulating propagation depth in predictor obtains the best performance. 
As the GIN encoder has highly non-linearity across layers, truncation would lead to non-explainability and somehow perturb the black-box. While the operation of relation graph in predictor updates node embedding in a linear way \eqref{eq:node}, adapting the propagation depth is harmonious with message passing process.

\begin{figure}[ht]
\centering
\subfigure[Without adaptation.\label{fig:tsne-a}]{
	\includegraphics[width=0.23\textwidth]{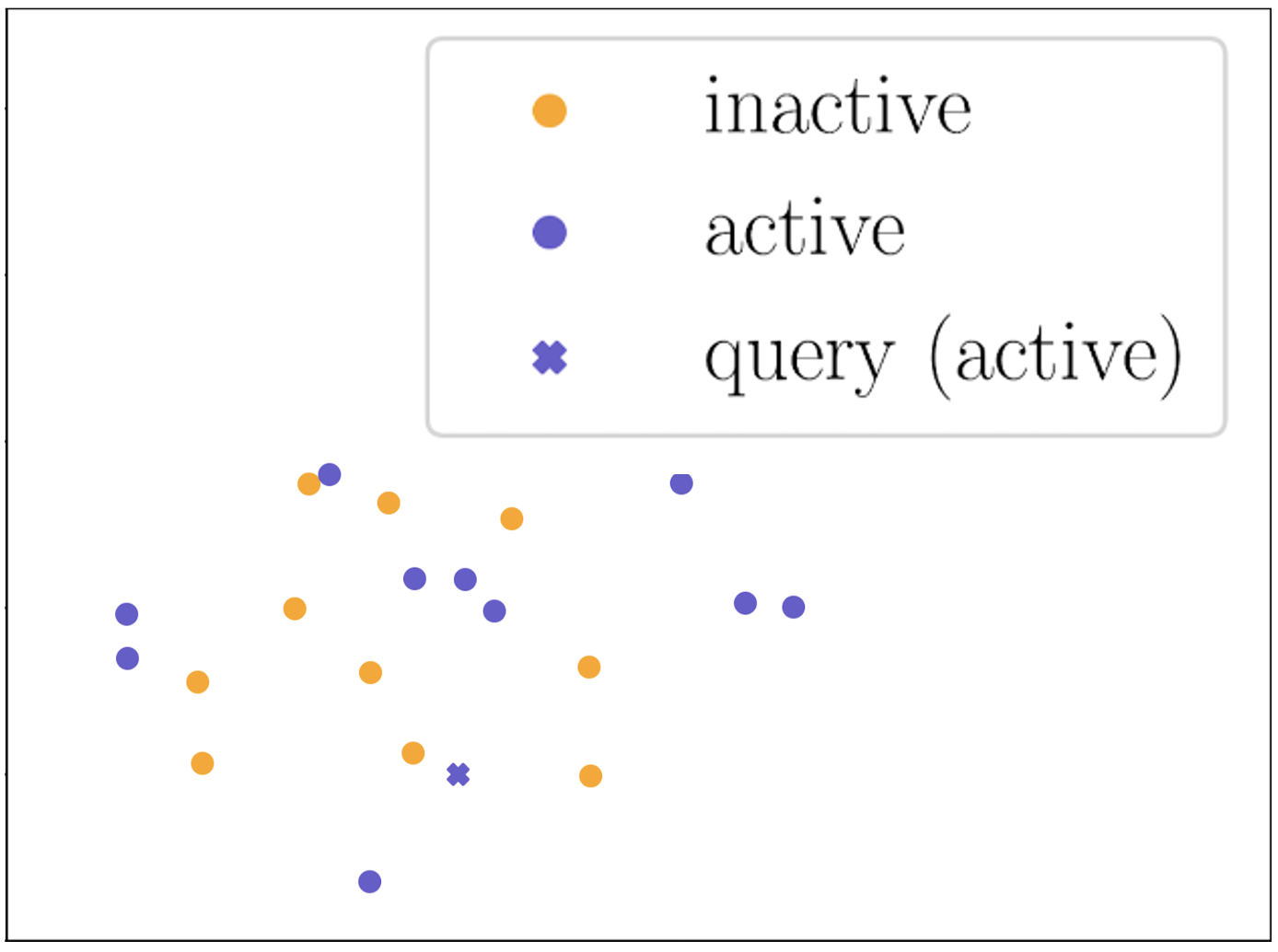}}
\subfigure[task-level adaptated.\label{fig:tsne-b}]{
	\includegraphics[width=0.23\textwidth]{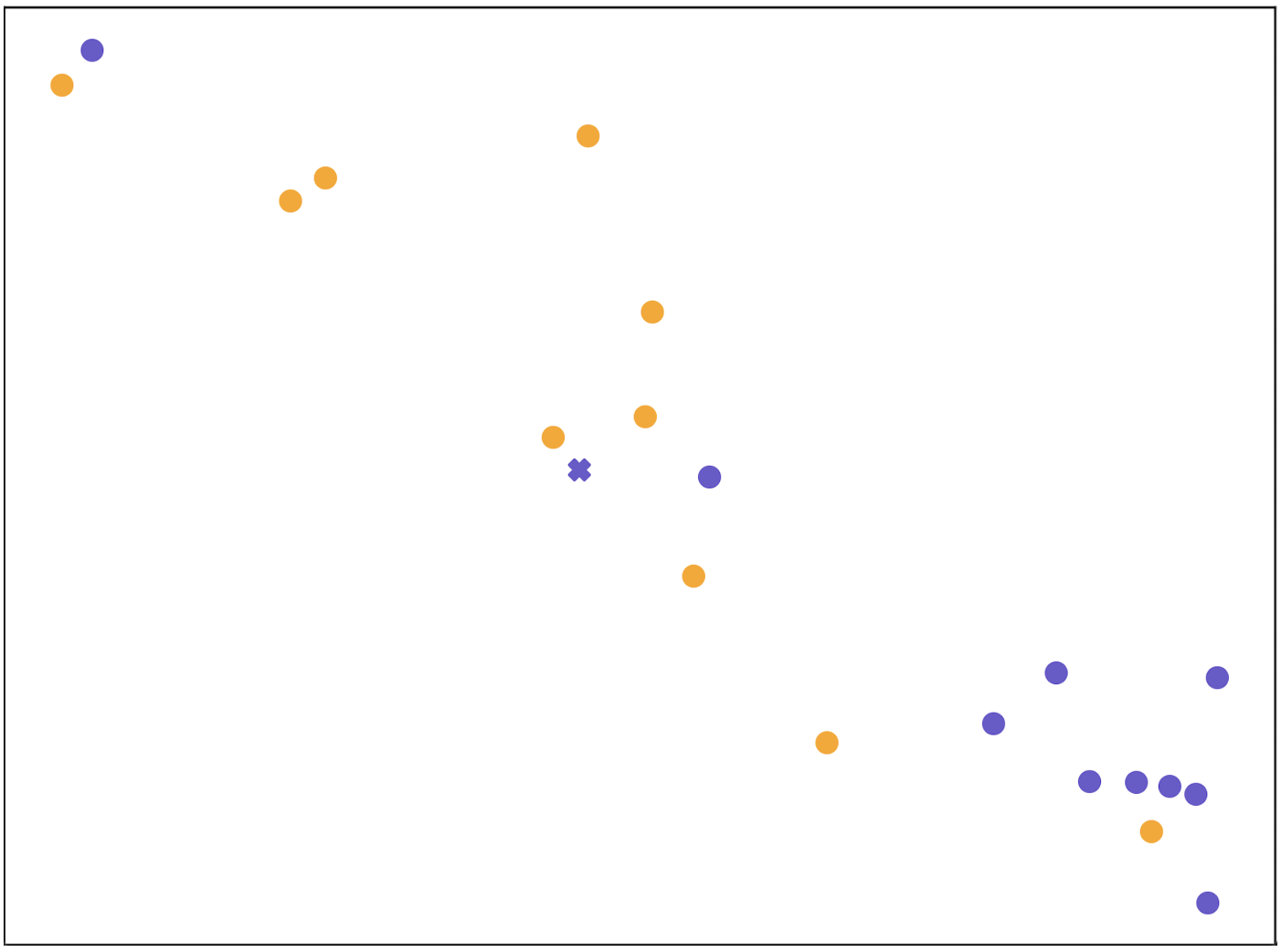}}
\subfigure[With relation graph.\label{fig:tsne-c}]{
	\includegraphics[width=0.23\textwidth]{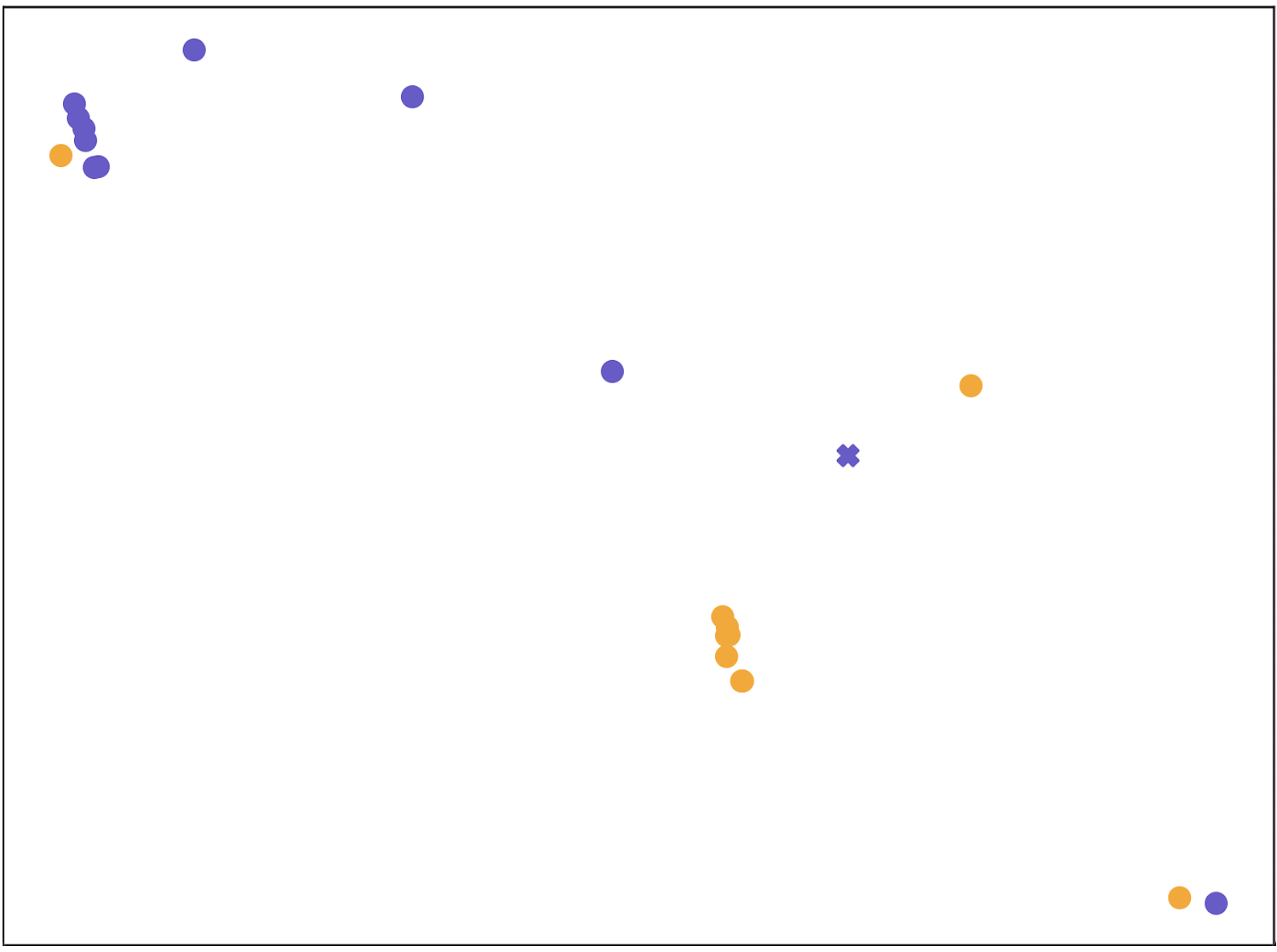}}
\subfigure[query-level adaptated.\label{fig:tsne-d}]{
	\includegraphics[width=0.23\textwidth]{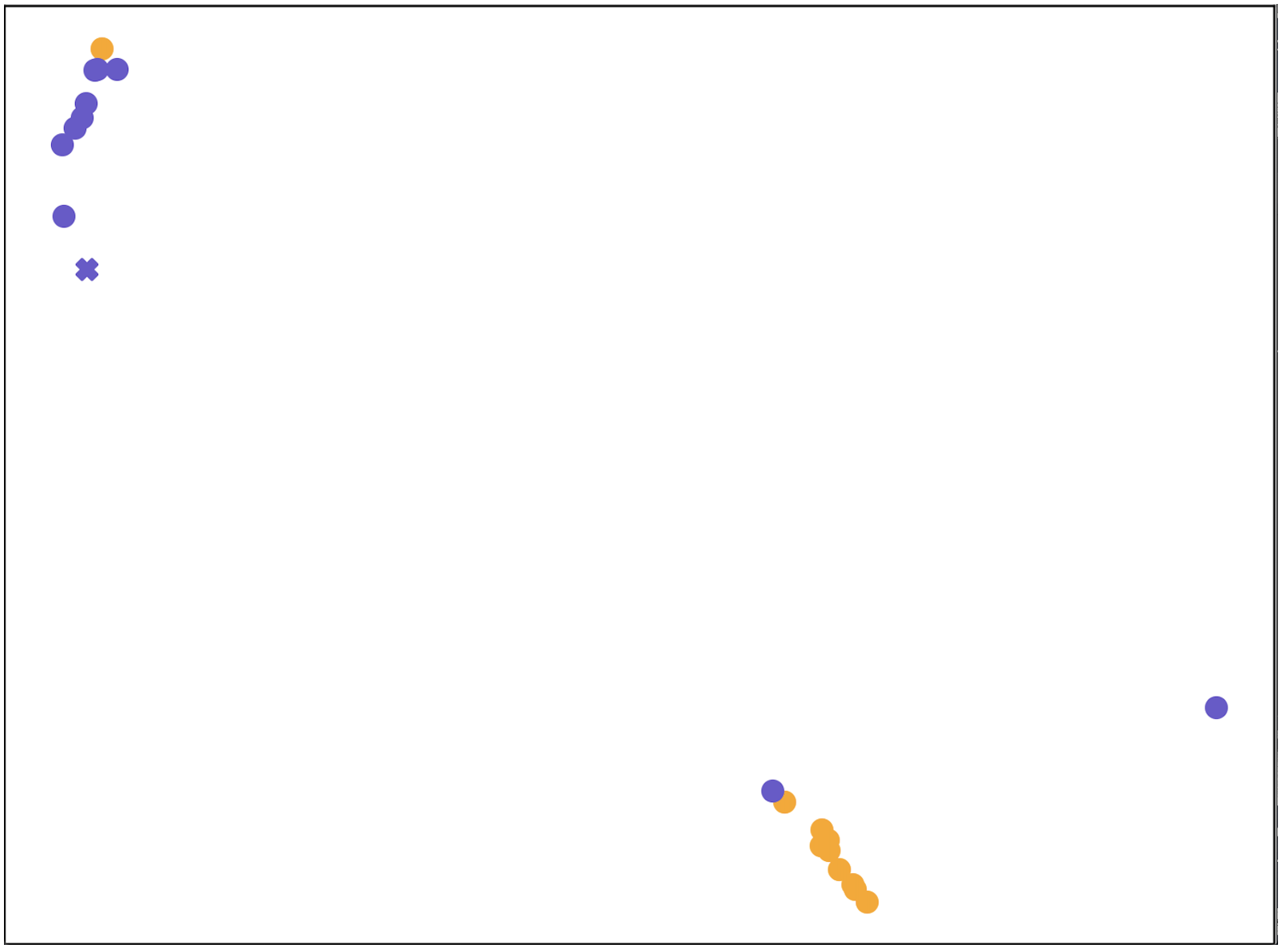}}
\caption{Molecular representation visualization for 10-shot case in task SR-p53 of Tox21. 
}
\label{fig:tsne}
\end{figure}
Figure~\ref{fig:tsne} shows the t-SNE visualization~\cite{van2008visualizing} of molecular representations learned on a 10-shot support set and a query molecule with ground-truth label ``active" in task SR-p53 from Tox21. 
As shown, 
molecular representations obtained without adaptation (Figure~\ref{fig:tsne-a}) are mixed up, 
since the encoder has not been adapted to the target property of the task. 
Molecular representations being processed by our property-adaptive encoder 
(Figure \ref{fig:tsne-b}) becomes more distinguishable, indicating that adapting molecular representation in task-level takes effect. 
Molecular representations in 
Figure~\ref{fig:tsne-c} 
and Figure~\ref{fig:tsne-d} form clear clusters as we encourage similar molecules to be connected 
during relation graph refinement by \eqref{eq:node}. 
The difference is that molecular representations in Figure~\ref{fig:tsne-c} are refined by the best propagation depth number for all tasks in 10-shot case, while 
molecular representations in Figure~\ref{fig:tsne-d} are refined by 4 depths of propagation which are selected for the specific query molecule. 
As shown, we can conclude that our molecular-adaptive refinement steps help better separate molecules of different classes. 
Ablation study of configurations of hypernetworks is in 
Appendix \ref{app:abl-hypernet}. 

\subsection{Study of Hierarchical Adaptation Mechanism}

\paragraph{Task-Level Adaptation.}
In \TheName{}, parameter-efficient task-level adaptation is achieved by  using hypernetworks to modulate the node embeddings during message passing. 
We compared this amortization-based adaptation with gradient-based adaptation in PAR which has similar main network  with \TheName{}. 
We record their adaptation process, i.e., time required to process the support set and the test performance. 
Table \ref{tab:adapt-new} shows the results. 
PAR uses molecules in support set to take gradient steps, and updates all parameters in GNN. 
We record each of a maximum five steps, where we can find that it easily overfits as the testing ROC-AUC keeps dropping with more steps. The time consumption also grows. 
In contrast, 
\TheName{} processes molecules in support set by hypernetworks, which is much more efficient as only one single forward pass is needed. \TheName{} can obtain better performance due to the reduction of adaptive parameters, which also leads to better generalization and alleviates the risk of overfitting to a few shots. Table \ref{tab:adapt-new} and Figure \ref{fig:model}(a) both indicate that the underlying overfitting problem can be mitigated by \TheName{}.

\begin{table}[ht]
	\center
	\setlength\tabcolsep{1.5pt}
	\begin{tabular}{c|c|c|c|c|c|c}
		\hline
		& \TheName{} & \multicolumn{4}{c}{PAR} \\\hline
		\# Total para.&3.28M& \multicolumn{4}{c}{2.31M}\\
		\# Adaptive para.&3.00K& \multicolumn{4}{c}{0.38M}\\\cline{2-7}
		\# Gradient steps &          -              & 1    & 2    & 3&4    & 5   \\
		ROC-AUC (\%) &        84.26            &    82.07  &   81.85  &   80.32  & 79.09 & 77.25  \\
		Time (secs)       &          1.09        &   2.02 &3.62& 5.34 &6.76  &8.10  \\\hline
	\end{tabular}
	\caption{Comparison of task-level adaptation approaches.}
	\label{tab:adapt-new}
\end{table}



\begin{figure}[H]
\centering
\includegraphics[width=1\columnwidth]{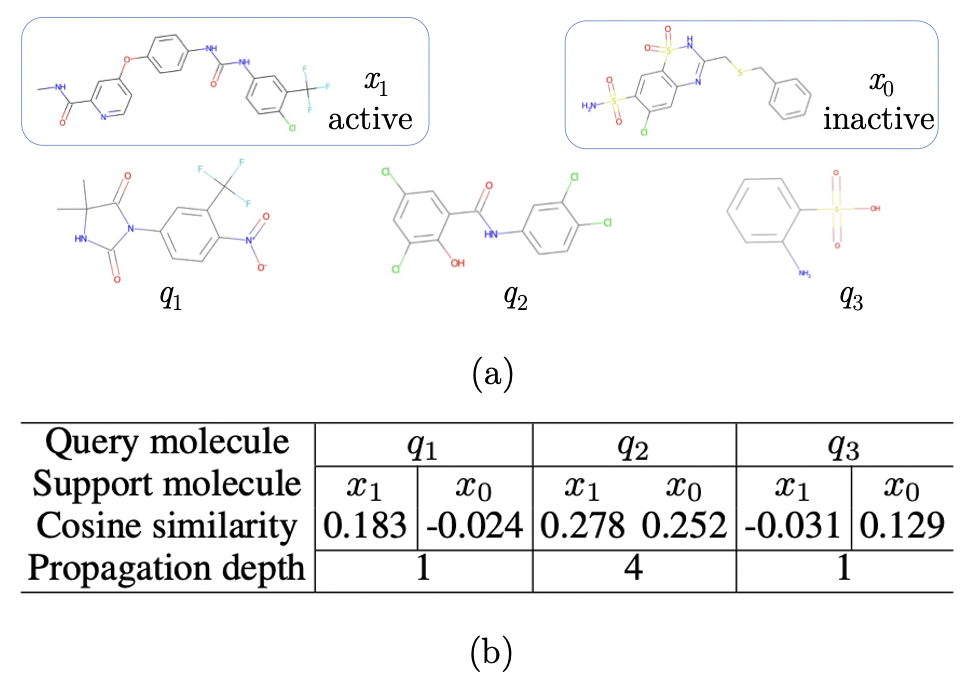}
\caption{Illustration of query-level adaptation. (a), Molecular graphs of support molecules $x_1,~x_0$ and query molecules $q_1,~q_2,~q_3$.
	(b), Cosine similarities between query molecules and support molecules, and propagation depth taken to classify each query molecule. 
}
\label{fig:molediff}
\end{figure}

\paragraph{Query-Level Adaptation.}
Finally, we present a case study on query-level adaptation. 
More experimental results on validating the design of query-level adaptation are in Appendix \ref{app:mole-adapt}.
We use a 1-shot support set and 3 query molecules in task SR-p53 of Tox21.
In Figure \ref{fig:molediff}(a), 
$x_1$ and $x_0$ are support molecules with different labels, $q_1$, $q_2$ and $q_3$ are query molecules. 
As shown, classifying $q_1$ and $q_3$ is relatively easy and the propagation depth will be 1, 
while classifying $q_2$ is hard and requires 4 depth of propagation.
Considering the shared substructures (function groups), 
$q_1$ and $x_1$ are visually similar, $q_3$ and $x_0$ are visually similar.  
While both $x_1$ and $x_0$ share substructures with $q_2$, it is hard to tell which of them is more similar to $q_2$. 
Figure \ref{fig:molediff}(a) provides the cosine similarity based on the molecule representations generated by Pre-GNN , 
which confirms our observation: $q_1$ is much more similar to $x_1$, $q_3$ is much more similar to $x_0$, and $q_2$ shows large similarities with both samples. 
Intuitively, classifying $q_1$ and $q_3$ will be easier  while $q_2$ will be hard. 
In the dynamic propagation of \TheName{}, we find different depths are taken: 
1 for both
$q_1$ and $q_3$ while 4 for $q_2$. \TheName{} achieves effective query-level adaptation by assigning more complex models for query molecules that are difficult to classify.

\section{Conclusion}
We propose \TheName{} to handle few-shot MPP in a parameter-efficient manner. 
We investigate two key factors in few-shot molecular property prediction under the common encoder-predictor framework: adaptation-efficiency and query-level adaptation. Evidence shows that too much adaptive parameter would lead to overfitting, thus we design a parameter-efficient GNN adapter, which can modulate node embedding and propagation depth of message passing of GNN in a unified way. 
We also notice the importance of capturing query-level difference and therefore propose hierarchical adaptation mechanism, which is achieved by using a unified GNN adapter in both encoder and predictor. Empirical results show that \TheName{} achieves the best performance on both MoleculeNet and FS-Mol.

\section*{Acknowledgments}

Q. Yao is supported 
by research fund of National Natural Science Foundation of China (No. 92270106), 
and Independent Research Plan of the Department
of Electronic Engineering Department at Tsinghua University.

\bibliographystyle{named}
\bibliography{fsmpp}

\clearpage
\appendix
\onecolumn

\section{Adopting MAML for Task-Level Adaptation}\label{app:maml}
Denote all model parameters as $\bm{\Theta}$. 
The model first predicts samples in support set and gets loss to do local-update. 
Denote the loss for local-update as
\begin{align*}
	\mL_{\tau}^S(\bm{\Theta})=\sum\nolimits_{\cX_{{\tau},s}\in\cS_{\tau}}\bm{y}_{{\tau},s}^\top \log
	\left( 
	\hat{\bm{y}}_{{\tau},s}
	\right), 
\end{align*}
where $\hat{\bm{y}}_{{\tau},s}$ is the prediction made by the main network with parameter $\bm{\Theta}$. 
The loss for global-update is calculated with samples in query set, which is obtained as
\begin{align*}
	\mL_{\tau}^Q(\bm{\Theta}'_{\tau})=\sum\nolimits_{\cX_{{\tau},q}\in\cQ_{\tau}}\bm{y}_{{\tau},q}^\top \log
	\left( 
	\hat{\bm{y}}_{{\tau},q}
	\right), 
\end{align*}
where $\hat{\bm{y}}_{{\tau},q}$ is the prediction made by the main network with parameter $\bm{\Theta}'_{\tau}$. 
Algorithm \ref{alg:maml-train} shows the meta-training procedure, and Algorithm \ref{alg:maml-test} shows the meta-testing procedure.
\begin{algorithm}[H]
	\caption{Meta-training procedure of MAML.}
	\label{alg:maml-train}
	\begin{algorithmic}[1]
		\REQUIRE meta-training task set $\bm{\cT}_{\text{train}}$
		\STATE initialize $\bm{\Theta}$ randomly;
		\WHILE{not done}
		\FOR {each task $\cT_{\tau} \in \bm{\cT}_{\text{train}}$}
		\STATE evaluate $\nabla_{\bm{\Theta}}\mL_{\tau}^S(\bm{\Theta})$ with respect to all samples in $\cS_{\tau}$;
		\STATE compute adapted parameters with gradient descent: $\bm{\Theta}'_{\tau}=\bm{\Theta}-\nabla_{\bm{\Theta}}\mL_{\tau}^S(\bm{\Theta})$;
		\ENDFOR
		\STATE update $\bm{\Theta}\leftarrow\bm{\Theta}-\nabla_{\bm{\Theta}}\sum\nolimits_{\cT_{\tau} \in \bm{\cT}_{\text{train}}}\mL_{\tau}^Q(\bm{\Theta}'_{\tau})$;
		\ENDWHILE
		\RETURN learned $\bm{\Theta}^*$.
	\end{algorithmic}
\end{algorithm}

\begin{algorithm}[H]
	\caption{Meta-testing procedure of MAML.}
	\label{alg:maml-test}
	\begin{algorithmic}[1]
		\REQUIRE learned $\bm{\Theta}*$, a meta-testing task $\cT_{\tau}$;
		\STATE evaluate $\nabla_{\bm{\Theta}}\mL_{\tau}^S(\bm{\Theta})$ with respect to all samples in $\cS_{\tau}$;
		\STATE compute adapted parameters with gradient descent: $\bm{\Theta}'_{\tau}=\bm{\Theta}-\nabla_{\bm{\Theta}}\mL_{\tau}^S(\bm{\Theta})$;
		\STATE make prediction $\bm{y}_{{\tau},q}$ for $\cX_{{\tau},q}\in\cQ_{\tau}$ with adapted parameter $\bm{\Theta}'_{\tau}$;
	\end{algorithmic}
\end{algorithm}

\section{More Details of \TheName{}}
\label{app:method}

\subsection{Encoder}
\label{app:gin}

\paragraph{Encoder for MoleculeNet.}
We use GIN~\cite{xu2019powerful}, a powerful GNN structure, as the encoder to encode molecular graphs. 
Each node embedding $\bh_v$ represents an atom, and each edge $e_{vu}$ represents a chemical bond.
In GIN, $\agg(\cdot)$ function in \eqref{eq:gnn-update} adds all neighbors up, while $\update(\cdot)$ adds the aggregated embeddings and the target node embedding and feeds it to a MLP:
\begin{align}
	\label{eq:gin-update}
	\bh_{v}^{l}
	=\texttt{MLP}^{l}_G
	\left( 
	(1+\epsilon)\bh_{v}^{l-1}+
	\sum\nolimits_{u\in\mathcal{H}(v)} \bh_{\text{u}}^{l - 1}
	\right),
\end{align}
where $\epsilon$ is a scalar parameter to distinguish the target node.
To obtain the molecular representation, $\readout(\cdot)$ function in \eqref{eq:gnn-readout} is specified as
\begin{align}
	\label{eq:gin-readout}
	\br=\texttt{MLP}_R
	\left(
	\text{MEAN}
	( 
	\{\bh_v^{L}| v\in\cV\}
	)\right).
\end{align}
\paragraph{Encoder for FS-Mol.}
Following existing works \cite{chen2022meta,schimunek2023context}, we directly adopt the PNA \cite{corso2020principal} network provided in FS-Mol benchmark \cite{stanley2021fs} as the molecular encoder.

\subsection{Predictor}
\label{app:classifier}
The classifier needs to make prediction of the query $\hat\bh_{{\tau},q}$, 
given $N_\tau$ labeled support samples $\{(\bh_{{\tau},s},y_{{\tau},s})~|~\cX_{{\tau},s}\in \cS_{\tau}\}$.
We adopt an adaptive classifier~\cite{requeima2019fast}, which maps the labeled samples in each class to the parameters of a linear classifier. 
For the active class, its classifier is obtained as 
\begin{align}
	\label{eq:adc-adaw}
	\bm{w}_{+}
	\! = \!
	\texttt{MLP}_{w+}
	\Big(
	\frac{1}{|\cS_{\tau}^+| }
	\!\!
	\sum\nolimits_{\cX_{{\tau},s} \in \cS_{\tau}^+}\bh_{{\tau},s}
	\Big),
	\\\label{eq:adc-adab}
	b_{+}
	\! = \!
	\texttt{MLP}_{b+}
	\Big(
	\frac{1}{|\cS_{\tau}^+| }
	\!\!
	\sum\nolimits_{\cX_{{\tau},s} \in \cS_{\tau}^+}\bh_{{\tau},s}
	\Big), 
\end{align}
where $\bm{w}_{+}$ has the same dimension with $\bh_{{\tau},q}$, and $b_+$ is a scalar. 
Likewise, the classifier for the inactive class is obtained as
\begin{align}
	\label{eq:adc-adaw2}
	\bm{w}_{-}
	\! = \!
	\texttt{MLP}_{w-}
	\Big(
	\frac{1}{|\cS_{\tau}^-| }
	\!\!
	\sum\nolimits_{\cX_{{\tau},s} \in \cS_{\tau}^-}\bh_{{\tau},s}
	\Big),
	\\\label{eq:adc-adab2}
	b_{-}
	\! = \!
	\texttt{MLP}_{b-}
	\Big(
	\frac{1}{|\cS_{\tau}^-| }
	\!\!
	\sum\nolimits_{\cX_{{\tau},s} \in \cS_{\tau}^-}\bh_{{\tau},s}
	\Big), 
\end{align}
Then the prediction is made by
\begin{align}
	\label{eq:adc-pred}
	\hat{\bm{y}}_{{\tau},q}=\texttt{softmax}([\bm{w}_-^{\top}\bh_{{\tau},q}+b_-,~\bm{w}_+^{\top}\bh_{{\tau},q}+b_+]),
\end{align}
where $\texttt{softmax}(\bm{x}) = \text{exp}(\bx) / \sum\nolimits_{i}\text{exp}\left(  [\bm{x}]_i  \right)$
and 
$[\bm{x}]_{i}$ means the $i$th element of $\bm{x}$.  

\subsection{Unified GNN Adapter}
The choice of modulation function $e(\cdot)$ in \eqref{eq:film} can be varying \cite{wu2023coldnas}, 
while in this work we adopt a simple feature-wise linear modulation (FiLM)~\cite{perez2018film} function.  

\subsection{Meta-Testing Procedure}\label{app:test}
Algorithm~\ref{alg:test} provides the testing procedure.
The procedure is similar to Algorithm~\ref{alg:train}.  
The node embeddings $\bh^{l'}$ are adapted by \eqref{eq:film}. The propagation depth is adapted by selecting the layer with the maximal plausibility using \eqref{eq:select} (line 7 and 16). 
Then, $\bh^{l'}$ are fed forward to classifier as \eqref{eq:linear-predictor}.
\begin{algorithm}[ht]
	\caption{Meta-testing procedure of \TheName{}.}
	\begin{algorithmic}[1]
		\REQUIRE learned $\bm{\Theta}^*$, a meta-testing task $\cT_{\tau}$;
		\FOR {each task $\cT_{\tau} \in \bm{\cT}_{\text{train}}$}
		\FOR {$l\in\{1,2,\cdots,L_{\text{enc}}\}$}
		\STATE + generate $[\bm{\gamma}^l_\tau,p^l_\tau]$ by \eqref{eq:ada-pt};
		\STATE modulate atom embedding ${\bh}^{l}_v\leftarrow e(\bh^l_v,\bm{\gamma}^l_\tau)$; 
		\STATE * update atom embedding $\bh_{v}^{l}$ by \eqref{eq:gnn-update};
		\ENDFOR
		\STATE select propagation depth by \eqref{eq:select} and obtain atom embedding after message passing $\bh^{L_{\text{enc}}}_v\leftarrow\bh^{l'}_v$;
		\ENDFOR
		\STATE *obtain molecular embedding   by \eqref{eq:gnn-readout};
		\FOR {each query $( {\cX}_{{\tau},q},  {y}_{{\tau},q})\in\cQ_{\tau}$}
		\FOR {$l\in\{1,2,\cdots,L_{\text{rel}}\}$}
		\STATE + generate $[\bm{\gamma}^l_{\tau,q},p^l_{\tau,q}]$s by \eqref{eq:ada-pq};
		\STATE  modulate molecular embedding  $\bh_{\tau,i}^{l}\leftarrow e(\bh_{\tau,i}^{l}\bm{\gamma}^l_{\tau,q})$; 
		\STATE * update molecular embeddings by \eqref{eq:adj}-\eqref{eq:node};
		\ENDFOR
		\STATE select propagation depth by \eqref{eq:select} and obtain molecular embedding after message passing	$\bh^{L_{\text{rel}}}_{\tau,i}\leftarrow\bh^{l'}_{\tau,i}$;
		\STATE obtain prediction  $\hat{\by}_{\tau,q}$ by \eqref{eq:linear-predictor}. 
		\ENDFOR
	\end{algorithmic}
	\label{alg:test}
\end{algorithm}

\subsection{Hyperparameters}
\label{app:structure}

\paragraph{Hyperparameters on MoleculeNet.}
The maximum layer number of the GNN $L_{\text{enc}}=5$, the maximum depth of the relation graph $L_{\text{rel}}=5$, 
During training, for each layer in GNN, we set dropout rate as 0.5 operated between the graph operation and FiLM layer. The dropout rate of $\text{MLP}$ in \eqref{eq:mean-rep} \eqref{eq:ada-pt} and \eqref{eq:ada-pq} is 0.1. For all baselines, we use Adam optimizer
with learning rate 0.006 and the maximum episode number is 25000. 
In each episode, the meta-training tasks are learned one-by-one, query set size $M=16$. 
The ROC-AUC is evaluated every 10 epochs on meta-testing tasks and the best performance is reported. 
Table~\ref{tab:model-detail} shows the details of the other parts.

\paragraph{Hyperparameters on FS-Mol.}
The maximum layer number of the GNN $L_{\text{enc}}=8$, the maximum depth of the relation graph $L_{\text{rel}}=5$, 
During training, the dropout rate of $\text{MLP}_L$ is 0.1. We use Adam optimizer
with learning rate 0.0001 and the maximum episode number is 3000. 
In each episode, the meta-training tasks are learned with batch size 16, support set size $N_\tau=64$, and the others are used as queries. 
The average precision s evaluated every 50 epochs on validation tasks and the the model with best validation performance is tested and reported. 
Table~\ref{tab:model-detail-fsmol} shows the details of the other parts. 
\begin{table*}[htbp]
	\setlength\tabcolsep{0.pt}
	\centering
	\resizebox{1\textwidth}{!}{%
		\begin{tabular}{c|c|c}
			\hline
			& Layers  & Output dimension \\ \hline
			\multirow{2}{*}{$\texttt{MLP}$ in \eqref{eq:mean-rep}} &  input $\frac{1}{| \cV_{{\tau},s} |}
			\sum\nolimits_{v\in\cX_{{\tau},s}} 
			\big[
			\bh_v^{l}
			~|~
			\bm{y}_{{\tau},s}
			\big]$, fully connected, LeakyReLU   &300         \\ 
			& 2$\times$fully connected with with residual skip connection,$\frac{1}{K}[
			\sum\nolimits_{\cX_{{\tau},s} \in \cS_{\tau}^+}(\cdot)~|~\sum\nolimits_{\cX_{{\tau},s} \in \cS_{\tau}^-}(\cdot)] $&300\\  \hline
			\texttt{MLP} in \eqref{eq:ada-pt}& 3$\times$fully connected with residual skip connection           &   601         \\ \hline
			\texttt{MLP} in \eqref{eq:ada-pq}& 3$\times$fully connected with residual skip connection           &   257         \\ \hline
			\multirow{2}{*}{$\texttt{MLP}_G^l$ in \eqref{eq:gin-update}} &  input $(1+\epsilon)\bh_{v}^{l-1}+
			\bh_{\text{agg}}^{l - 1}$, fully connected, ReLU             &  600              \\
			&  fully connected           &  300           \\ \hline
			\multirow{2}{*}{$\texttt{MLP}_R$ in \eqref{eq:gin-readout}} &  input $\readout
			\left( \{\bh_v^{T}| v\in\cV_{t,i}\}\right)$, fully connected, LeakyReLU             &  128              \\
			&  fully connected           &  128           \\ \hline
			\multirow{3}{*}{$\texttt{MLP}$ in \eqref{eq:adj}} &  input $\text{exp}(|\bh_{{\tau},i}^{l-1}-\bh_{{\tau},i}^{l-1}|)$, fully connected, LeakyReLU             &  256              \\
			&  fully connected, LeakyReLU           &  128           \\
			&  fully connected           &  1           \\\hline
			\multirow{2}{*}{$\texttt{MLP}$ in \eqref{eq:node}} &   fully connected, LeakyReLU             &  256              \\
			&  fully connected, LeakyReLU           &  128           \\\hline
			\multirow{3}{*}{$\texttt{MLP}_{w-}$ in \eqref{eq:adc-adaw}} &  input $\frac{1}{K}\sum_{y_{{\tau},s}=c}\bm{h}_{{\tau},s}$, fully connected with residual skip connection, LeakyReLU             &  128              \\
			&  2$\times$ (fully connected with residual skip connection, LeakyReLU)             &  128              \\
			&  fully connected     &  128           \\\hline
			\multirow{3}{*}{$\texttt{MLP}_{b-}$ in \eqref{eq:adc-adab}} &  input $\frac{1}{K}\sum_{y_{{\tau},s}=c}\bm{h}_{{\tau},s}$, fully connected with residual skip connection, LeakyReLU             &  128              \\
			&  2$\times$ (fully connected with residual skip connection, LeakyReLU)             &  128              \\
			&  fully connected     &  1          \\
			\hline
			\multirow{3}{*}{$\texttt{MLP}_{w+}$ in \eqref{eq:adc-adaw2}} &  input $\frac{1}{K}\sum_{y_{{\tau},s}=c}\bm{h}_{{\tau},s}$, fully connected with residual skip connection, LeakyReLU             &  128              \\
			&  2$\times$ (fully connected with residual skip connection, LeakyReLU)             &  128              \\
			&  fully connected     &  128           \\\hline
			\multirow{3}{*}{$\texttt{MLP}_{b+}$ in \eqref{eq:adc-adab2}} &  input $\frac{1}{K}\sum_{y_{{\tau},s}=c}\bm{h}_{{\tau},s}$, fully connected with residual skip connection, LeakyReLU             &  128              \\
			&  2$\times$ (fully connected with residual skip connection, LeakyReLU)             &  128              \\
			&  fully connected     &  1          \\
			\hline
	\end{tabular}}
	\caption{Details of model structure for MoleculeNet.}
	\label{tab:model-detail}
\end{table*}
\begin{table*}[htbp]
	\setlength\tabcolsep{0.pt}
	\centering
	\resizebox{1\textwidth}{!}{%
		\begin{tabular}{c|c|c}
			\hline
			& Layers  & Output dimension \\ \hline
			\multirow{2}{*}{$\texttt{MLP}$ in \eqref{eq:mean-rep}} &  input $\frac{1}{| \cV_{{\tau},s} |}
			\sum\nolimits_{v\in\cX_{{\tau},s}} 
			\big[
			\bh_v^{l}
			~|~
			\bm{y}_{{\tau},s}
			\big],$, fully connected, LeakyReLU   &512         \\ 
			&2$\times$ fully connected with with residual skip connection,$[\frac{1}{|\cS_{\tau}^+|}
			\sum\nolimits_{\cX_{{\tau},s} \in \cS_{\tau}^+}(\cdot)~|~\frac{1}{|\cS_{\tau}^-|}\sum\nolimits_{\cX_{{\tau},s} \in \cS_{\tau}^-}(\cdot)] $&512\\  \hline
			\texttt{MLP} in \eqref{eq:ada-pt}& 3$\times$fully connected with residual skip connection           &   1025         \\ \hline
			\texttt{MLP} in \eqref{eq:ada-pq}& 3$\times$fully connected with residual skip connection           &   513         \\ \hline
			\multirow{3}{*}{$\texttt{MLP}$ in \eqref{eq:adj}} &  input $\text{exp}(|\bh_{{\tau},i}^{l-1}-\bh_{{\tau},i}^{l-1}|)$, fully connected, LeakyReLU             &  256              \\
			&  fully connected, LeakyReLU           &  128           \\
			&  fully connected           &  1           \\\hline
			\multirow{2}{*}{$\texttt{MLP}$ in \eqref{eq:node}} &  fully connected, LeakyReLU             &  256              \\
			&  fully connected, LeakyReLU           &  256           \\\hline
			\multirow{3}{*}{$\texttt{MLP}_{w-}$ in \eqref{eq:adc-adaw}} &  input $\frac{1}{|\cS_{\tau}^\pm|}\sum_{y_{{\tau},s}=c}\bm{h}_{{\tau},s}$, fully connected with residual skip connection, LeakyReLU             &  256              \\
			&  2$\times$ (fully connected with residual skip connection, LeakyReLU)             &  256              \\
			&  fully connected     &  256           \\\hline
			\multirow{3}{*}{$\texttt{MLP}_{b-}$ in \eqref{eq:adc-adab}} &  input $\frac{1}{|\cS_{\tau}^\pm|}\sum_{y_{{\tau},s}=c}\bm{h}_{{\tau},s}$, fully connected with residual skip connection, LeakyReLU             &  256              \\
			&  2$\times$ (fully connected with residual skip connection, LeakyReLU)             &  256              \\
			&  fully connected     &  1          \\
			\hline
			\multirow{3}{*}{$\texttt{MLP}_{w+}$ in \eqref{eq:adc-adaw2}} &  input $\frac{1}{|\cS_{\tau}^\pm|}\sum_{y_{{\tau},s}=c}\bm{h}_{{\tau},s}$, fully connected with residual skip connection, LeakyReLU             &  256              \\
			&  2$\times$ (fully connected with residual skip connection, LeakyReLU)             &  256              \\
			&  fully connected     &  256           \\\hline
			\multirow{3}{*}{$\texttt{MLP}_{b+}$ in \eqref{eq:adc-adab2}} &  input $\frac{1}{|\cS_{\tau}^\pm|}\sum_{y_{{\tau},s}=c}\bm{h}_{{\tau},s}$, fully connected with residual skip connection, LeakyReLU             &  256              \\
			&  2$\times$ (fully connected with residual skip connection, LeakyReLU)             &  256              \\
			&  fully connected     &  1          \\
			\hline
	\end{tabular}}
\caption{Details of model structure for FS-Mol.}
	\label{tab:model-detail-fsmol}
\end{table*}

\section{Comparison with Existing Few-Shot MPP Methods}
\label{app:compare}

We compare the proposed \TheName{} with existing few-shot MPP approaches in Table~\ref{tab:comp}. 
As shown, like MHNfs, \TheName{} also achieves the following properties: support of pretraining, task-level adaptation, query-level adaptation and fast-adaptation. It is noteworthy that query-level adaptation in MHNfs is archived with additional data, while  that in \TheName{} is achieved with meta-learned hypernetwork.
With the help of hypernetworks, 
our method can adapt at task-level and query-level more effectively and efficiently, without additional data.

\begin{table*}[ht]
	\centering
	\setlength\tabcolsep{7pt}
	\begin{tabular}{c|c|c|c | c| c}
		\hline
		Method    &   Support    & \multicolumn{2}{c|}{Hierarchical adaptation} & Fast       & Adaptation \\
		& pretraining & task-level &       query-level       & adaptation & strategy             \\ \hline
		IterRefLSTM &  $\times$  &    $\surd$     &          $\times$           &    $\surd$        & Pair-wise similarity  \\ 
		Meta-MGNN  & $\surd$    &    $\surd$     &          $\times$           &       $\times$      & Gradient              \\ 
		PAR     &$\surd$  &    $\surd$     &          $\times$           &   $\times$          & Attention+Gradient    \\ 
		ADKF-IFT     &$\surd$  &    $\surd$     &          $\times$           &   $\times$          & Gradient+statistical learning    \\ 
		MHNfs     &$\surd$  &    $\surd$     &          $\surd$           &   $\surd$          & Attention+pair-wise similarity    \\ 
		GS-META     &$\surd$  &    $\times$     &          $\surd$           &   $\surd$          & Message passing   \\ 
		\TheName{}     &  $\surd$ &    $\surd$     &           $\surd$           &      $\surd$      & Hypernetwork      \\ \hline
	\end{tabular}
\caption{Comparison of the proposed \TheName{} with existing few-shot MPP methods.}
	\label{tab:comp}
\end{table*}

\section{More Details of Experiments}
\label{app:exp-detail}
Experiments are conducted on a 24GB NVIDIA GeForce RTX 3090 GPU, with Python 3.8.13, CUDA version 11.7, Torch version 1.10.1.

\subsection{Datasets}
\label{app:dataset}

\paragraph{MoleculeNet.}
There are four sub-datasets for few-shot MPP:  Tox21~\cite{Tox21}, SIDER~\cite{kuhn2016sider}, MUV~\cite{rohrer2009maximum} and ToxCast~\cite{richard2016toxcast}, which are included in MoleculeNet~\cite{wu2018moleculenet}. We 
adopt the task splits provided by existing works \cite{altae2017low,wang2021property}. 
Tox21 is a collection of nuclear receptor assays related to human toxicity, containing 8014 compounds in 12 tasks, among which 9 are split for training and 3 are split for testing.
SIDER collects information about side effects of marketed medicines, and it contains 1427 compounds in 21 tasks, among which 21 are split for training and 6 are split for testing.
MUV contains compounds designed to be challenging for virtual screening for 17 assays, containing 93127 compounds in 17 tasks, among which 12 are split for training and 5 are split for testing.
ToxCast collects compounds with toxicity labels, containing 8615 compounds in 617 tasks, among which 450 are split for training and 167 are split for testing.

\paragraph{FS-Mol.}
FS-Mol benchmark, which contains a set of few-shot learning tasks for molecular property prediction carefully collected from ChEMBL27 \cite{mendez2019chembl} by \cite{stanley2021fs}. Following existing works \cite{chen2022meta,schimunek2023context}, we use the same 10\% of all tasks which contains 233,786 unique compounds, split into training (4,938 tasks), validation (40 tasks), and test (157 tasks) sets. Each task is associated with a protein target.

\subsection{Baselines}\label{app:baselines}
We compare our method with following baselines: 
\begin{itemize}[leftmargin=*]
	\item 
	\textbf{GNN-ST}\footnote{\url{https://github.com/microsoft/FS-Mol/}}~\cite{stanley2021fs}: A GNN as encoder and a MLP as predictor are trained from scratch for each task, using the support set. A GIN~\cite{xu2019powerful} is adopted on MoleculeNet. A PNA~\cite{corso2020principal} is adopted on FS-Mol.
	
	\item
	\textbf{MAT}\footnote{\url{https://github.com/microsoft/FS-Mol/}}~\cite{maziarka2020molecule}: Molecule Attention Transformer is adopted as encoder.
	
	\item 
	\textbf{GNN-MT}\footnote{\url{https://github.com/microsoft/FS-Mol/}}~\cite{stanley2021fs}: A task-shared GNN and task-specific MLPs are trained in a multi-task learning with all data in meta-training set. A GIN~\cite{xu2019powerful} is adopted on MoleculeNet. A PNA~\cite{corso2020principal} is adopted on FS-Mol.
	
	\item 
	\textbf{ProtoNet}\footnote{\url{https://github.com/jakesnell/prototypical-networks}}~\cite{snell2017prototypical}: It makes classification according to inner-product similarity between the target and the prototype of each class. This method is incorporated as a classifier after the GNN encoder.
	\item 
	\textbf{MAML}\footnote{\url{https://github.com/learnables/learn2learn}}~\cite{finn2017model}: It learns a parameter initialization and the model is adapted to each task via a few gradient steps w.r.t. the support set. We adopt this method for all parameters in a model composed of a GNN encoder and a linear classifier.
	\item 
	\textbf{Siamese}~\cite{koch2015siamese}: It learns two neural networks which are symmetric on structure to identity whether the input molecule pairs are from the same class. The performance is copied from~\cite{altae2017low} due to the lack of code.

	\item 
	\textbf{EGNN}\footnote{\url{https://github.com/khy0809/fewshot-egnn}}~\cite{kim2019edge}: It builds a relation graph that samples are refined, and it learns to predict edge-labels in the relation graph. This method is incorporated as the predictor after the GNN encoder. 
	
	\item 
	\textbf{IterRefLSTM}~\cite{altae2017low}: It  introduces matching networks combined with long short-term memory (LSTM) to refine the molecular representations according to the task context. The performance is copied from~\cite{altae2017low} due to the lack of code.
	\item 
	\textbf{PAR}\footnote{\url{https://github.com/tata1661/PAR-NeurIPS21}}~\cite{wang2021property}: It introduces an attention mechanism to capture task-dependent property and an inductive relation graph between samples, and incorporates MAML to train.
	\item 
	\textbf{ADKF-IFT}\footnote{\url{https://github.com/Wenlin-Chen/ADKF-IFT}}~\cite{wang2021property}: It uses gradient-based strategy to learn the encoder where it proposes Implicit Function Theory to avoid computing the hyper-gradient. And a Gaussian Process is learned from scratch in each task as classifier.
\end{itemize}	

\subsection{Performance Comparison on MoleculeNet with Pretrained Encoders}
\label{app:performance-pretrain}
\paragraph{Baselines with Pretrained Encoders.}
Here, 
we equip our \TheName{} with a pretrained encoder, and name it as 
\textbf{Pre-\TheName{}}.
We compare it with baselines with (w/) pretrained encoders. 
For fair comparison, all baselines use the same pretrained encoder~\cite{hu2019strategies}, which pretrained on ZINC15 dataset~\cite{sterling2015zinc}.  
The compared baselines include: 
\begin{itemize}[leftmargin=*]
\item 
\textbf{Pre-GNN}\footnote{\url{http://snap.stanford.edu/gnn-pretrain}}~\cite{hu2019strategies}: It trains a GNN encoder on ZINC15 dataset with graph-level and node-level self-supervised tasks, and fine-tunes the pretrained GNN on downstream tasks. We adopt the pretrained GNN encoder and a linear classifier.
\item 
\textbf{GraphLoG}\footnote{\url{http://proceedings.mlr.press/v139/xu21g/xu21g-supp.zip}}~\cite{xu2021self}: It introduces hierarchical prototypes to capture the global semantic clusters. And adopts an
online expectation-maximization algorithm to learn. We adopt the pretrained GNN encoder and a linear classifier.
\item 
\textbf{MGSSL}\footnote{\url{https://github.com/zaixizhang/MGSSL}}~\cite{hu2019strategies}: It trains a GNN encoder on ZINC15 dataset with graph-level, node-level and motif-level self-supervised tasks, and fine-tunes the pretrained GNN on downstream tasks. We adopt the pretrained GNN encoder and a linear classifier.
\item 
\textbf{GraphMAE}\footnote{\url{https://github.com/THUDM/GraphMAE}}~\cite{hou2022graphmae}: It presents a masked graph autoencoder for generative self-supervised graph pretraining and focus on feature reconstruction with both a masking strategy
and scaled cosine error. We adopt the pretrained GNN encoder and a linear classifier.
\item 
\textbf{Meta-MGNN}\footnote{\url{https://github.com/zhichunguo/Meta-Meta-MGNN}}~\cite{guo2021few}: It incorporates self-supervised tasks such as bond reconstruction and atom type prediction to be jointly optimized via MAML. It uses the pretrained GNN encoder provided by~\cite{hu2019strategies}.
\item 
\textbf{Pre-PAR}: The same as PAR but uses the pretrained GNN encoder provided by~\cite{hu2019strategies}.
\item 
\textbf{Pre-ADKF-IFT}: The same as ADKF-IFT but uses the pretrained GNN encoder provided by~\cite{hu2019strategies}.
\end{itemize}


\begin{table*}[ht]
	\centering
	\setlength\tabcolsep{1.5pt}
	\begin{tabular}{c|cc|cc|cc|cc}
		\hline
		\multirow{2}{*}{Method}  &\multicolumn{2}{c|}{Tox21} &\multicolumn{2}{c|}{SIDER}
		& \multicolumn{2}{c|}{MUV}&\multicolumn{2}{c}{ToxCast} \\
		&10-shot&1-shot&10-shot&1-shot&10-shot&1-shot&10-shot&1-shot\\\hline
		Pre-GNN&$83.02_{(0.13)}$&$82.75_{(0.09)}$&$77.55_{(0.14)}$&$67.34_{(0.30)}$&$67.22_{(2.16)}$&$65.79_{(1.68)}$&$73.03_{(0.67)}$&$71.26_{(0.85)}$\\
		GraphLoG&$81.61_{(0.35)}$&$79.23_{(0.93)}$&$75.18_{(0.27)}$&$67.52_{(1.40)}$&$67.83_{(1.65)}$&$66.56_{(1.46)}$&$73.92_{(0.15)}$&$73.10_{(0.39)}$\\
		MGSSL&$83.24_{(0.09)}$&$\underline{83.21}_{(0.12)}$&$77.87_{(0.18)}$&$69.66_{(0.21)}$&$68.58_{(1.32)}$&$66.93_{(1.74)}$&$73.51_{(0.45)}$&$72.89_{(0.63)}$\\
		GraphMAE&$84.01_{(0.27)}$&$81.54_{(0.18)}$&$76.07_{(0.15)}$&$67.60_{(0.38)}$&$67.99_{(1.28)}$&$\underline{67.50}_{(2.12)}$&$74.15_{(0.33)}$&$72.67_{(0.71)}$\\
		Meta-MGNN&$83.44_{(0.14)}$&$82.67_{(0.20)}$&${77.84}_{(0.34)}$&$74.62_{(0.41)}$&${68.31}_{(3.06)}$&$66.10_{(3.98)}$&$74.69_{(0.57)}$&$73.29_{(0.85)}$\\
		Pre-PAR&${84.95}_{(0.24)}$&$\underline{83.01}_{(0.28)}$&$\underline{78.05}_{(0.15)}$&$\underline{75.29}_{(0.32)}$&$69.88_{(1.57)}$&$66.96_{(2.63)}$&${75.48}_{(0.99)}$&$\underline{73.90}_{(1.21)}$\\
		Pre-ADKF-IFT&$\underline{86.06}_{(0.35)}$&$80.97_{(0.48)}$&$70.95_{(0.60)}$&$62.16_{(1.03)}$&$\bm{95.74}_{(0.37)}$&$67.25_{(3.87)}$&$\bm{76.22}_{(0.13)}$&$71.13_{(1.15)}$\\
		Pre-\TheName{}&$\bm{86.40_{(0.27)}}$&$\bm{84.35_{(0.14)}}$&$\bm{83.97_{(0.22)}}$&$\bm{80.70_{(0.28)}}$&$\underline{73.43_{(1.96)}}$&$\bm{69.26_{(2.35)}}$&$\bm{76.22_{(0.73)}}$&$\bm{75.09_{(0.95)}}$ \\\hline
	\end{tabular}
\caption{Test ROC-AUC obtained with pretrained GNN encoder. }
	\label{tab:results-pre}
\end{table*}

\paragraph{Performance with Pretrained Encoders.}
Table~\ref{tab:results-pre} shows the results. 
As shown, Pre-\TheName{} obtains significantly better performance except the 10-shot case on MUV, 
surpassing the second-best method Pre-ADKF-IFT by 3.10\%. 
MGSSL defeats the other methods which fine-tune pretrained GNN encoders, 
i.e., Pre-GNN, GraphLoG, and GraphMAE.
However, it still performs worse than Pre-\TheName{}, 
which validates the necessity of designing a few-shot MPP method instead of simply fine-tuning a pretrained  GNN encoder.
Moreover, comparing Pre-\TheName{} and \TheName{} in Table~\ref{tab:results}, the pretrained encoder brings 3.05\% improvement in average performance due to a better starting point of learning.
{
	
\subsection{Performance Given More Training Samples}\label{app:more-shot}
\TheName{} can handle general MPP problems. We conduct experiments on SIDER to validate PACIA given increasing labeled samples per task.  
We compare with GIN \cite{xu2019powerful}, which is a powerful encoder to handle MPP problems. To make fair comparison, we adapt the conventional multi-task learning manner. We first train a task-shared GIN with task-specific binary classifier on all meta-training data, then inherit the GIN while use the support set of a meta-testing task to learn a task-specific classifier from scratch. 

Figure \ref{fig:more-shot} shows the results. 
As can be seen, PACIA outperforms GIN for $1, 10, 16, 32, 64$-shot tasks, and is on par with GIN for $128$-shot tasks.
The performance gain of PACIA is more significant when fewer labeled samples are provided. 
Note that all parameters of GNN are fine-tuned, while PACIA only uses a few adaptive parameters to modulate the message passing process of GNN. 
The empirical evidence shows that PACIA nicely achieves its goal: handling few-shot MPP problem in a parameter-efficient way. 

\begin{figure}[ht]
\centering
\includegraphics[width=0.40\columnwidth]{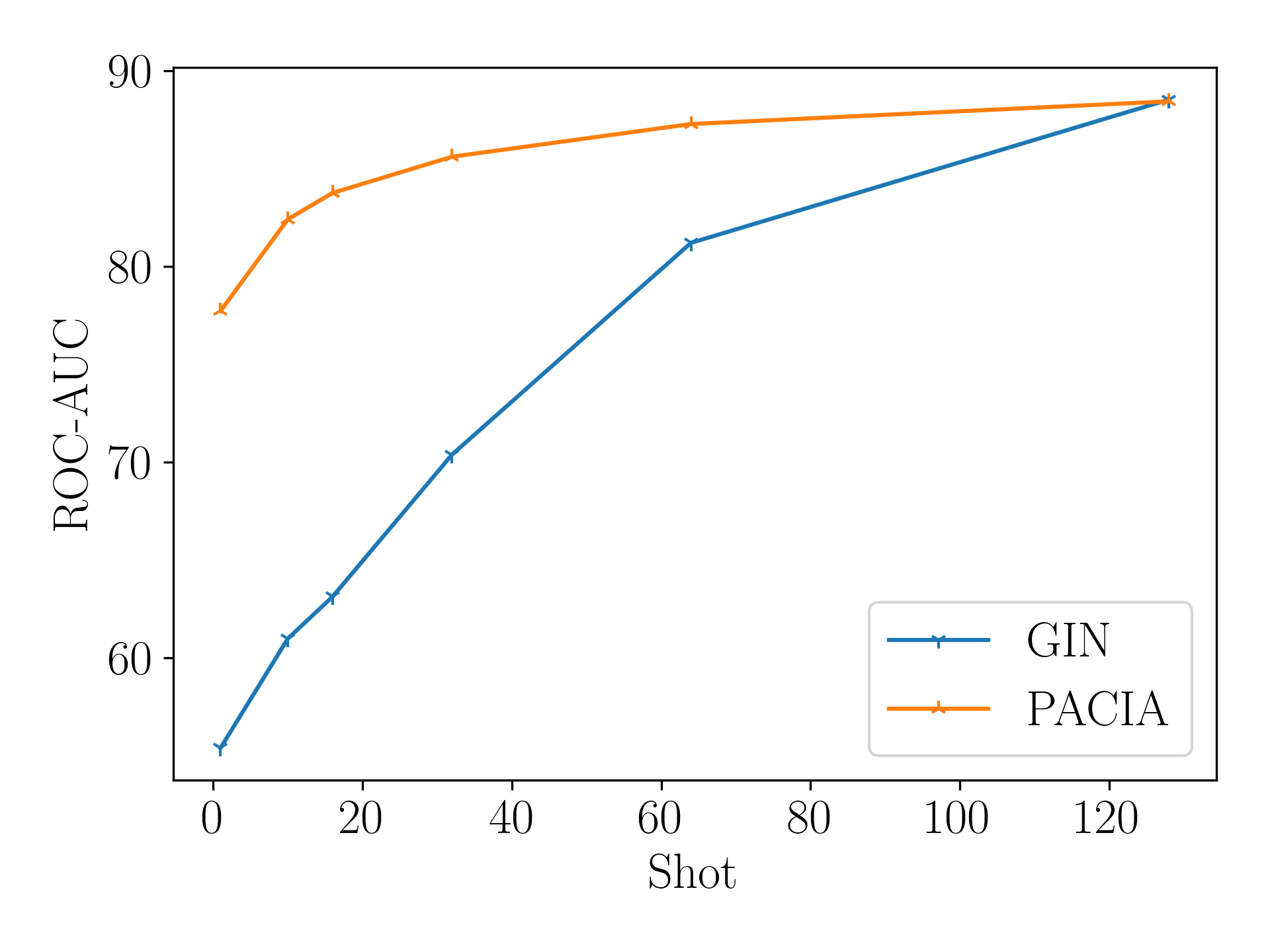}
\vspace{-10px}
\caption{Testing ROC-AUC of  \TheName{} and GIN on SIDER, with different number of labeled samples (shot) per task.}
\label{fig:more-shot}
\end{figure}

\subsection{Ablation Study of Hypernetworks}\label{app:abl-hypernet}
	
	For ablation studies, we provide results concerning with three aspects in hypernetworks:
	
	\paragraph{Effect of Concatenating Label.}
	We show effect of concatenating label $\bm{y}_{{\tau},s}$. Table \ref{tab:abl-hyper1} shows the testing ROC-AUC obtained on SIDER. As shown, "w/ Label" helps keeping the label information in support set, which improves the performance.
	
	\begin{table}[ht]
		\centering
		\setlength\tabcolsep{7pt}
		
		\begin{tabular}{c|c|c}
			\hline
			& 10-shot &       1-shot         \\ \hline
			w/ Label &  $82.40_{(0.26)}$  &   $77.72_{(0.34)}$  \\ 
			w/o Label  & $76.91_{(0.17)}$  &    $74.10_{(0.41)}$       \\ \hline
		\end{tabular}
	\caption{Effect of concatenating label $\bm{y}_{{\tau},s}$. Performance is evaluated by testing ROC-AUC (\%) on SIDER.}
		\label{tab:abl-hyper1}
		
	\end{table}
	
	\paragraph{Different Ways of Combining Prototypes.}
	We show performance with different ways of combining active prototype $\bm{r}_{{\tau},+}^l$ and inactive prototype $\bm{r}_{{\tau},-}^l$ in \eqref{eq:ada-pt} and \eqref{eq:ada-pq}. Table \ref{tab:abl-hyper2} show the results.
	As "Concatenating" active and inactive prototypes allows MLP to capture more complex patterns, it obtains better performance on SIDER as shown in Table \ref{tab:abl-hyper2}.
	
	\begin{table}[ht]
		\centering
		\setlength\tabcolsep{7pt}
		
		\begin{tabular}{c|c|c}
			\hline
			& 10-shot &       1-shot         \\ \hline
			Concatenating&  $82.40_{(0.26)}$  &   $77.72_{(0.34)}$  \\ 
			Mean-Pooling  & $79.67_{(0.23)}$   &   $75.08_{(0.29)}$    \\ \hline
		\end{tabular}
	\caption{Different ways of combining active prototype $\bm{r}_{{\tau},+}^l$ and inactive prototype $\bm{r}_{{\tau},-}^l$. Performance is evaluated by testing ROC-AUC (\%) on SIDER.}
		\label{tab:abl-hyper2}
	\end{table}
	
	\paragraph{Effect of Different MLP Layers.}
	We show performance with different layers of MLP in hypernetworks. 
	Table \ref{tab:abl-hyper3} shows the testing ROC-AUC obtained on SIDER. 
	Here, we constrain that the MLPs in \eqref{eq:mean-rep}\eqref{eq:ada-pt}\eqref{eq:ada-pq} have the same layer number.
	As shown, using 3 layers reaches the best performance. 
	Please note that although we can set different layer numbers for MLPs used in  \eqref{eq:mean-rep}\eqref{eq:ada-pt}\eqref{eq:ada-pq} which further improves performance, setting the same layer number already obtains the state-of-the-art performance. Hence, we set layer number as 3 consistently. 
	
	\begin{table}[ht]
		\centering
		\setlength\tabcolsep{2pt}
		\begin{tabular}{c|c|c|c|c}
			\hline
			& 1 layer &  2 layer &3 layer &4 layer       \\ \hline
			10-shot&  $79.98_{(0.35)}$ &  $81.85_{(0.33)}$&  $82.40_{(0.26)}$&   $82.43_{(0.28)}$ \\ 
			1-shot & $75.02_{(0.40)}$  &   $76.56_{(0.36)}$&   $77.72_{(0.34)}$&   $77.59_{(0.31)}$     \\ \hline
		\end{tabular}
	\caption{Effect of layers of MLP in hypernetworks. Performance is evaluated by testing ROC-AUC (\%) on SIDER.}
		\label{tab:abl-hyper3}
	\end{table}
}

\subsection{A Closer Look at Query-Level Adaptation}
\label{app:mole-adapt}
In this section, we pay a closer look at our query-level adaptation mechanism, proving evidence of its effectiveness.

\subsubsection{Performance under Different Propagation Depth}
\begin{figure}[ht]
	\centering
	\subfigure[10-shot case.]{
		\includegraphics[width=0.48\textwidth]{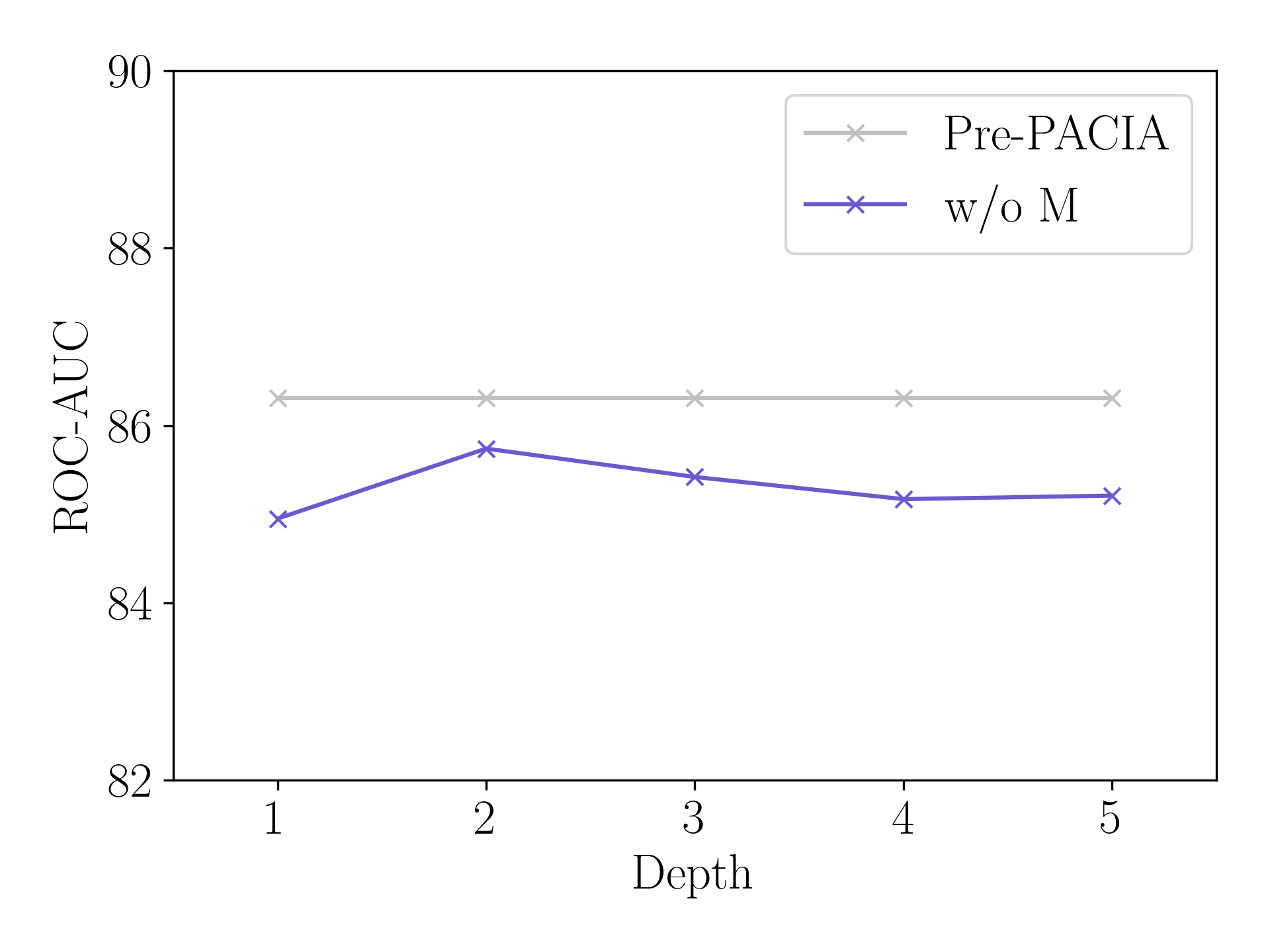}}
	\subfigure[1-shot case.]{
		\includegraphics[width=0.48\textwidth]{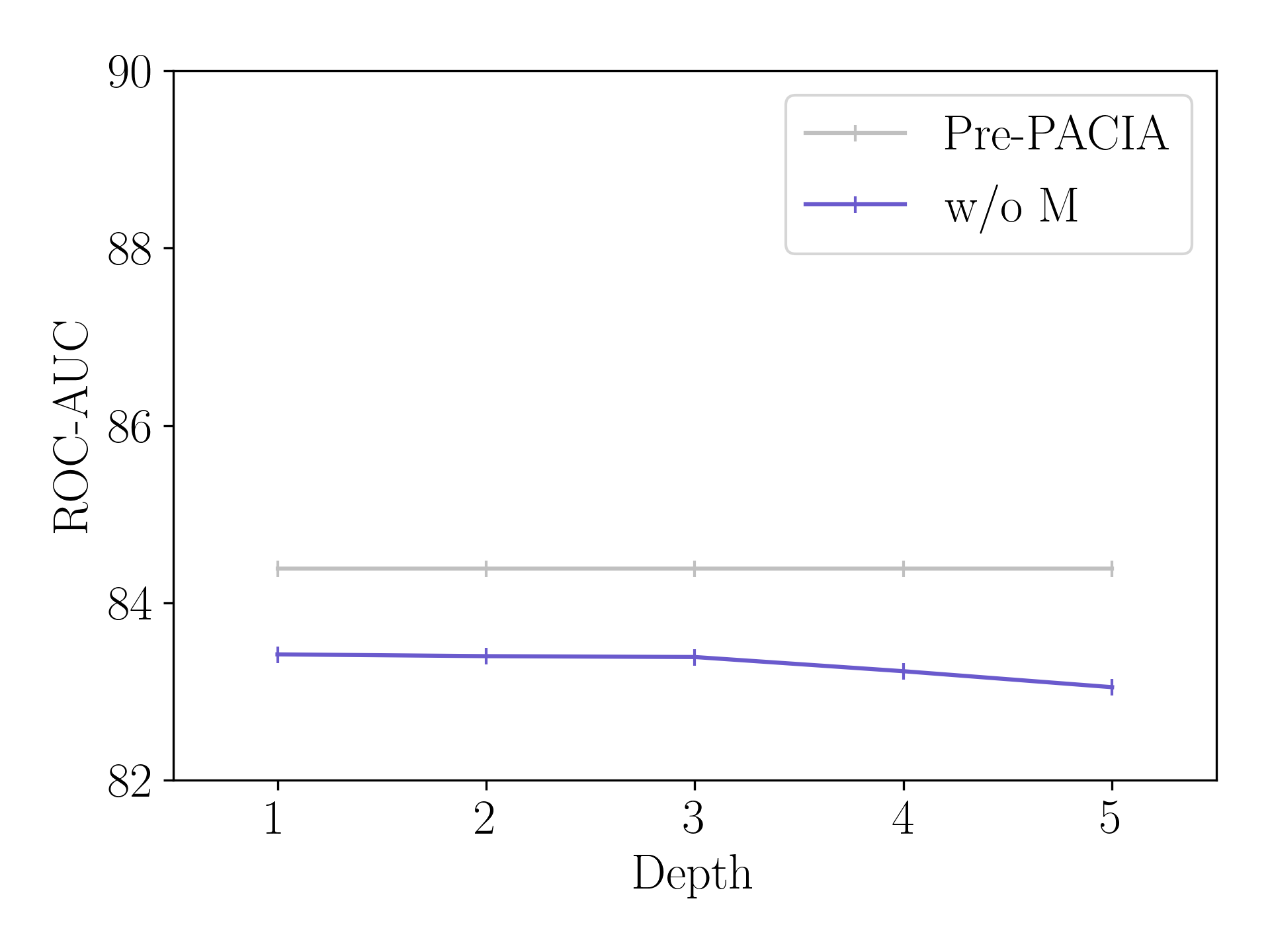}}
	\caption{Comparing Pre-\TheName{} with ``w/o M" using different fixed propagation depth of relation graph on Tox21.}
	\label{fig:layer-num}
\end{figure}

Figure \ref{fig:layer-num} compares Pre-\TheName{} with ``w/o M" (introduced in Section~\ref{sec:ablation}) using different fixed layers of relation graph refinement on Tox21, where the maximum depth $L=5$. 
As can be seen, 
Pre-\TheName{} equipped performs much better than 
``w/o M" which takes the same depth of relation graph refinement as in PAR. 
This validates the necessity of query-level adaptation. 

\subsubsection{Distribution of Propagation Depth}
\begin{figure}[H]
	\centering
	\subfigure[Distribution of learned $l'$.]
	{\includegraphics[width=0.48\columnwidth]{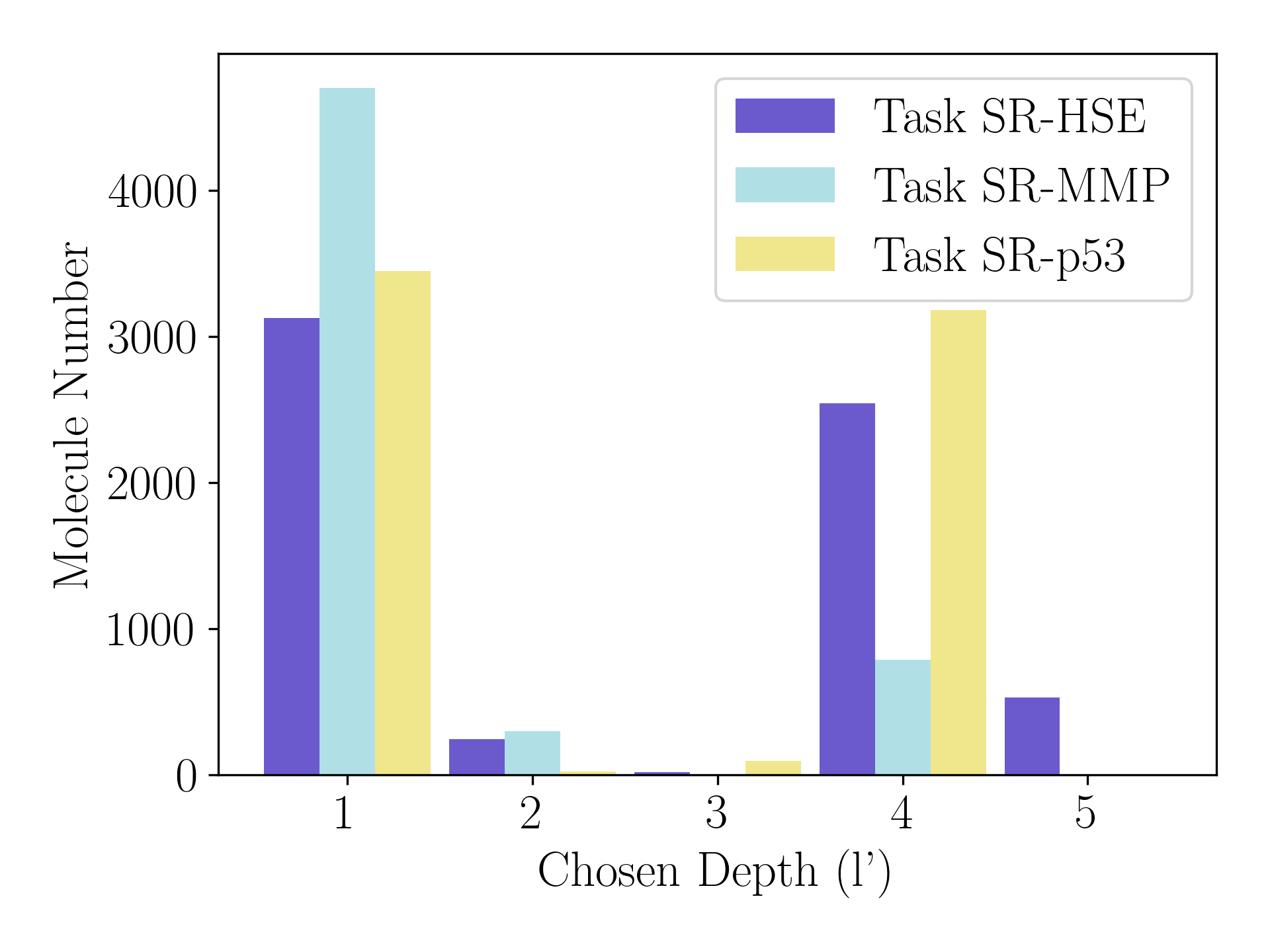}}
	\label{fig:moledistrib}
	\subfigure[Performance comparison.]{
		\includegraphics[width=0.48\columnwidth]{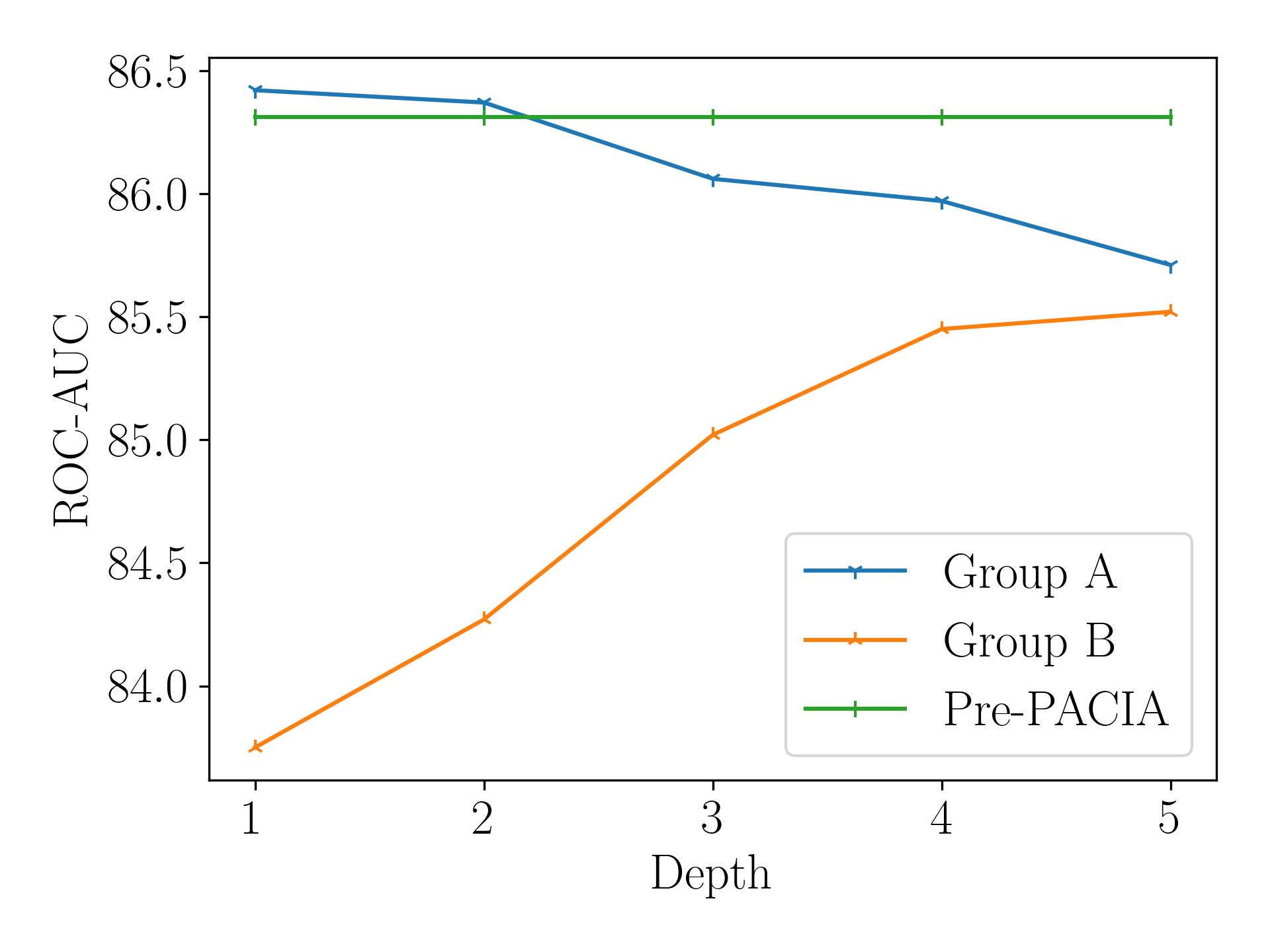}}
	\label{fig:groups}
	\caption{Examine query-level adaptation of Pre-\TheName{} on 10-shot tasks of Tox21. 
	}
	\label{fig:mada}
\end{figure}

Figure \ref{fig:mada}(a) plots the distribution of  learned $l'$ for query molecules in meta-testing tasks for 10-shot case of Tox21. 
The three meta-testing tasks contain different number of query molecules in scale: 6447 in task SR-HSE, 5790 in task SR-MMP, and 6754 in task SR-p53. 
We can see that Pre-\TheName{} choose different $l'$ for query molecules in the same task. 
Besides, the distribution of  learned $l'$ varies across different meta-testing tasks: 
molecules in task SR-MPP  mainly choose smaller depth while 
molecules in the other two tasks tend to choose greater depth. 
This can be explained as most molecules in task SR-MPP are relatively easy to classify,  which is consistent with the fact that Pre-\TheName{} obtains the highest ROC-AUC on SR-MPP among the three meta-testing tasks (83.75 for SR-HSE, 88.79 for SR-MPP and 86.39 for SR-p53). 

Further, we pick out molecules with $l'=1$ (denote as \textbf{Group A}) and $l'=4$ (denote as \textbf{Group B}) as they are more extreme cases. 
We then 
apply ``w/o M" with different fixed depth for Group A and  Group B, and compare them with Pre-\TheName{}. 
Figure \ref{fig:mada}(b) shows the results. 
Different observations can be made for these two groups.
Molecules in Group A have better performance with smaller depth relation graph, they can achieve higher ROC-AUC score than the average of all molecules using Pre-\TheName{}. 
These indicate they are easier to classify and it is reasonable that Pre-\TheName{} choose $l'=1$ for them. 
While molecules in Group B are harder to classify and require $l'=4$. 

{
}

\end{document}